# In Search of Meaning: Lessons, Resources and Next Steps for Computational Analysis of Financial Discourse

Mahmoud El-Haj[*]

Paul Rayson[*]

Martin Walker[#]

Steven Young[†]

Vasiliki Simaki[‡]

This draft: March 2019

[†] Corresponding author: Lancaster University Management School, Lancaster University, U.K., email: s.young@lancaster.ac.uk  [*] School of Computing and Communications, Lancaster University, U.K. [#] Alliance Manchester Business School, Manchester University, U.K. [‡] Department of Linguistics and English Language, and Centre for Corpus Approaches in Social Science (CASS), Lancaster University. We are grateful for comments and guidance from Peter Pope and Andrew Stark (editors), Daniel Bens, Ted Christiansen, Mahmoud Gad, Katherine Schipper (discussant), and Stephen Taylor. We are also grateful to Paulo Alves, José Costa, Andrew Moore and Steve Whatham Support for support with data and software resources described herein. Research support was provided by. Financial support was provided by the Economic and Social Research Council (ESRC) (contracts ES/J012394/1, ES/K002155/1, ES/R003904/1 and ES/S001778/1) and the Research Board of the Institute of Chartered Accountants in England and Wales. A preliminary draft of this paper was circulated under the title "Computational Analysis of Financial Narratives: Overview, Critique, Resources and Future Directions". Details of annual report winners and non-winner control firms used in Appendix B are available from the authors on request. Associated annual report content is available from public sources.

# In Search of Meaning: Lessons, Resources and Next Steps for Computational Analysis of Financial Discourse

**Abstract**

We critically assess mainstream accounting and finance research applying methods from computational linguistics (CL) to study financial discourse. We also review common themes and innovations in the literature and assess the incremental contributions of work applying CL methods over manual content analysis. Key conclusions emerging from our analysis are: (a) accounting and finance research is behind the curve in terms of CL methods generally and word sense disambiguation in particular; (b) implementation issues mean the proposed benefits of CL are often less pronounced than proponents suggest; (c) structural issues limit practical relevance; and (d) CL methods and high quality manual analysis represent complementary approaches to analyzing financial discourse. We describe four CL tools that have yet to gain traction in mainstream AF research but which we believe offer promising ways to enhance the study of meaning in financial discourse. The four tools are named entity recognition (NER), summarization, semantics and corpus linguistics.

# In Search of Meaning: Lessons, Resources and Next Steps for Computational Analysis of Financial Discourse

## 1. Introduction

Information is the lifeblood of financial markets and the amount of data available to decision-makers is increasing exponentially. Bank of England (2015) estimates that 90% of global information has been created during the last decade, the vast majority of which is unstructured data (e.g., free-form text).[1] The dramatic growth in written and spoken data is clearly evident in financial markets. For example, Dyer et al. (2017) find a 113% increase in the median length of U.S. registrants' 10-K annual reports over the period 1996-2013 and Lewis and Young (2019) report similar results for U.K. annual reports. For many applications, the volume of unstructured financial data exceeds the capacity of humans to process the content manually. Users of financial market data are therefore turning increasingly to computational linguistics to assist with the task of processing large volumes of unstructured data.[2] Academic research in accounting and finance is mirroring this trend.

Our paper has three objectives. First, we critically assess mainstream accounting and finance research that applies methods from computational linguistics (CL) to study written and spoken language (discourse) in financial markets. Our critique views extant research through the following three lenses: consistency with the core principles in CL; performance against the advantages of automated textual analysis proposed by Li (2010a); and practical relevance. Second, we review common themes and innovations in the literature and assess the incremental contributions of work applying CL methods over manual content analysis. Third, we describe a suite of CL tools that are yet to gain traction in mainstream accounting and finance research but which we believe offer promising ways to enhance the study of meaning in financial discourse.

---

[1] In 1998, Merrill Lynch projected that available data will expand to 40 zettabytes (one zettabyte equals one trillion gigabytes) by 2020 and estimated somewhere between 80-90% of all potentially usable business information may originate in unstructured form. Reinsel et al. (2018) forecast that the global datasphere will expand to 175 zettabytes by 2025. Although the estimate includes video and image data as well as structured data in databases, the majority compromises plain text.
[2] Throughout this paper we use "computational linguistics" as shorthand for the areas of natural language processing (NLP), text-focused artificial intelligence (AI) and information retrieval from computer science, plus the smaller group of empirical methods developed in the field of corpus linguistics involving frequency-based approaches to studying language.

A number of studies review aspects of CL research in accounting and finance (AF). Li (2010a) evaluates the benefits of CL methods over manual content analysis and reviews the first wave of studies using automated methods to examine accounting disclosures. Loughran and McDonald (2016) extend Li's (2010a) work by combining an updated review of the literature with a more focused survey and description of methods that characterize extant work in AF. In particular, they critique work on readability, highlight the importance of transparency when describing the process of converting raw text to quantitative measures, and reiterate Li's (2010a) call for economic theory to drive choice of CL methods rather than vice versa. Kearney and Liu (2014) narrow the focus further by reviewing work on sentiment analysis published in finance before 2013. Finally, Fisher et al. (2016) synthesize the stream of natural language processing (NLP) research utilizing AF data and identify paths for future research. Their review suggests a disconnect between mainstream AF research employing CL methods and the broader computer science literature using accounting and finance datasets.

Prior surveys start from the premise that the motivation for advocating CL analysis of financial text is compelling. This view is not universally accepted, however, although the basis for scepticism is not clearly articulated. We motivate our review by critiquing four reasons underpinning such suspicion: (a) doubt over the value of studying narrative disclosures; (b) distrust of CL approaches to scoring text; (c) cynicism about the validity of applying CL methods to financial market disclosures; and (d) concern over the way methods are applied and the relevance of the research questions examined. We conclude that the final explanation represents the most credible argument against using CL tools to analyse financial text. We proceed to evaluate research in light of this concern using three lenses.

Our first evaluation lens compares the application of methods in AF research to four core principles and practices that underpin the CL approach (corpus building, annotation, NLP, and evaluation). Our approach differs from previous surveys that define the textual analysis landscape according to the state-of-the-art in AF. We conclude that while beacons of best practice are evident, mainstream AF research appears to be behind the curve in terms of CL sophistication generally, and word sense disambiguation in particular, when judged against computer science and even specialist subfields within AF. Our second evaluation lens is the advantages of CL analysis (over manual coding) proposed by Li (2010a). Predicted benefits include lower scoring costs, wider generalizability, greater objectivity, improved replicability, enhanced statistical power, and scope for identifying "hidden" linguistic features. We conclude that these benefits are often less pronounced than AF research portrays. Our third evaluation lens assesses the relevance of extant work to debates in policy and practice regarding the role and value of financial discourse. We argue that relevance is constrained by at least two factors. First, the majority of CL research in AF operates at an aggregate level such as the entire 10-K or the complete Management Discussion and Analysis (MD&A), whereas practitioners, standard setters and regulators are often interested in more granular issues such as the format and content of specific disclosures, placement of content within the overall reporting package, limits on the use of jargon concerning particular topics, etc. Second, it is not immediately obvious how commonly employed empirical proxies for discourse quality such as readability (Fog index), tone (word-frequency counts) and text reuse (cosine similarity) map into the practical properties of effective communication identified by financial market regulators.

With these caveats in mind, we proceed to review common themes and innovations in the literature and assess the incremental contributions of work applying CL methods over manual content analysis. The median AF study examines 10-K filings using basic content analysis methods such as readability algorithms and keyword counts. The degree of clustering is consistent with the initial phase of the research lifecycle, with agendas shaped as much by ease of data access and implementation as by research priorities. Nevertheless, closer inspection reveals how relatively basic word-level methods have been used to provide richer insights into the properties and effects of financial discourse. Refinements to standard readability metrics, development of domain-specific wordlists, and the use of weighting schemes and text filtering to improve word-sense disambiguation represent welcome advances on naïve unigram word counts. We also acknowledge a move towards the use of more NLP technology in the form of machine learning and topic modeling, although the trend is characterized by a narrow methodological focus that lags best practice. We conclude that the main weakness of AF research is its continuing reliance on bag-of-words methods that fail to reflect context and meaning.

Analysis of financial discourse using manual content methods has a long tradition in AF (Merkl-Davies and Brennan 2007). Establishing incremental contribution for studies adopting a CL lens is not a straightforward task. A significant fraction of CL-focused studies appears to re-examine broadly similar issues to those previously explored using manual methods and arrives at broadly similar conclusions. We argue that emerging CL research should take greater care to ensure that the evidence from extant manual content analysis studies is afforded appropriate recognition. We also note that some important discourse-related research questions and settings in AF do not lend themselves naturally to the CL treatment. A key conclusion to emerge from our review is that CL methods and high-quality manual analysis represent symbiotic approaches to analyzing financial discourse. Both approaches are associated with comparative advantages and comparative weaknesses. The challenge for researchers choosing the CL route is to ensure that research questions align closely with the fundamental comparative advantages of scalability and latent feature detection.

In addition to taking a more critical and dispassionate approach to evaluating the contribution of automated textual analysis research in AF, we extend Li (2010a) and Loughran and McDonald (2016) by adopting a forward-looking perspective on CL methods and their applicability. Specifically, we review the following four tools from the CL field that offer significant potential for AF researchers: named entity recognition (NER), summarization, semantic analysis, and corpus methods. As well as helping to extend research horizons in financial discourse, the discussion speaks to the question posed by Loughran and McDonald (2016: 1223) regarding the potential benefits of parsing more deeply for contextual meaning in a business context. Their concern is that using more complex methods beyond simple word counts that ignore the sequence in which words are presented (i.e., meaning) may add more noise than signal to the empirical construct. We discuss tools from CL specifically designed to improve the signal-to-noise ratio by disambiguating word meaning.

## 2. Automated analysis of narrative disclosures in accounting and finance research

While qualitative disclosures have long attracted the interest of AF researchers, the need to hand collect and manually score content constrained work in this area. Early work by Abrahamson and Amir (1996), Antweiler and Frank (2004), Das and Chen (2007), Henry (2006, 2008), Tetlock (2007), Tetlock et al. (2008), Li (2008) and Loughran and McDonald (2011) paved the way for the recent upsurge in work applying CL methods to study the properties and consequences of qualitative information. Proponents view these methods as offering a cost effective solution to perceived weaknesses of manual content analysis including low replicability, limited generalizability, and low statistical power (Li 2010a). The adoption of automated scoring techniques in mainstream AF is helping to move research towards a more balanced treatment of quantitative and qualitative data. Scepticism nevertheless surrounds the merits of studying financial narratives using CL methods. Understanding these concerns is the first step to critiquing the contribution of this rapidly expanding body of research.

We propose four (non-mutually exclusive) reasons for scepticism about the benefits of studying financial discourse using CL methods. The first reason reflects fundamental doubt about the incremental value of qualitative disclosures over quantitative data. Summary quantitative measures such as asset prices, accounting earnings, and analysts' forecasts impound information from multiple sources including qualitative disclosures. Further, quantitative data are often considered more objective and verifiable. Collectively, these features could limit the relevance of qualitative information in financial markets. This view is nevertheless hard to reconcile with theory and evidence. Theory establishes a clear link from disclosure to efficient pricing and contracting (Beyer at al. 2010). Verrecchia (1983) demonstrates that precision determines information value in a pricing context and there is no inherent reason why discourse must lack precision. Empirically, the incremental usefulness of qualitative data has been established in multiple contexts such as: explaining contemporaneous stock returns (Francis et al. 2002, Asquith et al. 2005); predicting short-term stock returns (Tetlock 2007, Chen et al. 2014); predicting future earnings (Kothari et al. 2009, Feldman et al. 2010); predicting accounting violations (Goel et al. 2010, Larcker and Zakolyukina 2012), fraud (Purda and Skillicorn 2015) and bankruptcy (Smith and Taffler 2000); and reducing information asymmetry (Lang and Stice-Lawrence 2015). The irrelevance of unstructured data is not, therefore, a valid argument for dismissing analysis of financial discourse, regardless of the particular methodology employed.

A second reason to be sceptical about the value of CL methods reflects general distrust of the view that language is amenable to computerized analysis. Written text and spoken language represent highly sophisticated forms of communication where context is critical to interpreting meaning. Even in situations where context is clearly defined and understood by message sender and receiver, ambiguity in meaning is still commonplace (Navigli 2009). Accordingly, cynicism over the ability of algorithms to extract information and meaning reliably is perhaps not surprising when the human brain struggles to interpret signals correctly. Despite the undeniable challenges, CL research consistently demonstrates that it is possible to use algorithms and statistical procedures to measure the properties of text and extract meaning (Sparck Jones 2001). CL methods are now ubiquitous and underpin everyday activities from surfing the web, to using product reviews when shopping, to electing politicians. For researchers in AF to dismiss this body of work as either economically irrelevant or theoretically flawed is not credible.

A third reason to be suspicious of the CL approach is that the domain-specific features of financial discourse render it less amenable to automated processing than other corpora. A large proportion of CL work focuses on non-specialist corpora constructed from newspaper articles, product reviews, political speeches, social media posts and popular fiction. Generic, non-technical language free of jargon and idiosyncratic content is more amenable to automated processing by off-the-shelf tools and requires little domain expertise. In contrast, financial discourse is complex, jargon-heavy and context-specific. For example, the Fog index for the average 10-K annual report exceeds 19.0, implying the disclosure is unreadable according to standard interpretations of the measure (Li 2008).[3] Li (2010a) discusses the classification challenges posed by the lack of business-specific dictionaries; Loughran and McDonald (2011) demonstrate empirically that the Harvard-IV-4 TagNeg (H4N) dictionary has low classification accuracy for 10-Ks since almost three-quarters of the H4N list is not normally considered negative in a finance setting; and Loughran and McDonald (2016) discuss how standard readability metrics based on sentence length are less applicable to financial reporting than other forms of prose such as novels and political speeches. More generally, Loughran and McDonald (2016: 1223) question whether advanced NLP methods can be applied to business text.

While financial disclosures undoubtedly feature complex domain-specific content, these characteristics alone are not sufficient to reject the CL approach. Extant work finds that acceptable levels of classification accuracy are achievable in financial applications when appropriate methods and reasonable adjustments are applied (Loughran and McDonald 2011, Henry and Leone 2016, Fisher et al. 2016). Moreover, NLP methods are routinely applied in domains featuring similar or higher levels of technical content including medicine and healthcare (Olaronke and Olaleke 2015), education (Burstein et al. 2017), and biodiversity science (Thessen et al. 2012). Domain complexity is not, therefore, a reason on its own to dismiss CL approaches to analyzing financial discourse.

A fourth reason for suspicion is concern over the way CL methods are applied and the (ir)relevance of the research questions examined. Both Li (2010a) and Loughran and McDonald (2016) hint at these concerns in their respective reviews. For example, both studies argue that future research needs to focus more on economic fundamentals rather than seeking applications of off-the-shelf textual methods from CL. Further, Loughran and McDonald (2016) emphasize the importance of exposition and transparency in the next phase of AF research. While criticism of extant work is implicit in these calls to action, neither review critiques prior research directly on these dimensions. Accordingly, the validity or otherwise of this concern remains an open question that warrants further scrutiny.

---

[3] Consistent with this view, financial market regulators routinely raise concerns about complex disclosures and stress the need to simplify language (Securities and Exchange Commission 1998, Financial Reporting Council 2015).

We proceed to evaluate prior research in light of this concern using three complementary lenses. The first evaluation lens (section 3) compares the application of methods in AF research to discipline norms in CL. The second lens (section 4) evaluates extant work against the suite of advantages to CL analysis (over manual coding) highlighted by Li (2010a). The third lens (section 5) assesses the relevance of extant CL research in AF to policy and other non-academic debates regarding the role and value of financial discourse.

## 3. Assessment against core principles in computational linguistics

This section evaluates textual analysis research in AF against four principles of CL that collectively determine good practice, replicability and application of the scientific method. To help frame the discussion and the remainder of the paper, Appendix A summarizes key steps in the CL pipeline, provides an overview of broad approaches available to researchers wishing to analyze financial discourse, and illustrates how these approaches currently differ across AF and CL disciplines. The appendix provides a general framework for AF researchers new to the area.

3.1 Corpus creation

Arguably the most fundamental principle in CL is that research is based on evidence derived from real language examples. CL researchers therefore invest significant effort constructing corpora to support their experiments, with the aim of creating samples that are representative of a language or variety of language. These data are then released as open access to provide a common basis for further investigation.[4] In addition to building general corpora that are representative of the styles and sources for a particular language type (e.g. spoken and written, fiction and non-fiction, newspapers, books, letters, etc.), researchers also create corpora to study specific questions such as use of metaphors in palliative care (Semino et al. 2018) and construction of professional identities in corporate mission statements (Koller 2011).

Corpus construction as practiced in CL is not an activity that mainstream AF research has pursued. Since most studies focus on distinct sources such as annual reports, earnings press releases, conference calls, media articles, etc., construction of corpora that are representative of financial language in general is rare.[5] An exception is Kothari et al.'s (2009) analysis of text from three financial market sources: annual reports, analysts' reports, and the financial media. However, their approach does not follow standard corpus construction rules and even more critically the corpus is not released publicly. The same holds for the majority of extant AF studies that create a corpus from a single language source such as 10-K annual reports. El-Haj et al. (2019) is the only AF study of which we are aware that publishes its corpora of U.K. annual report sections for further analysis.

The absence of recognized financial market corpora restricts linguistic endeavour in the field by limiting replicability and incremental analysis. It also leaves open the question of whether it is valid to use wordlists developed from one source (e.g., 10-K filings) to measure the same linguistic feature in an alternative source (e.g., earnings announcements, conference calls, etc.). Davies and Tama-Sweet (2012), Dikolli et al. (2017) and Bushee et al. (2018) report systematic differences in linguistic features across different financial corpora, raising doubts about the practice of applying static wordlists to different language sources. In contrast. Loughran and McDonald (2011) provide evidence that their wordlists constructed using 10-K filings are transferrable to other financial domains. Accordingly, while evidence consistently indicates that domain-specific wordlists outperform general language dictionaries (Loughran and McDonald 2011, Henry and Leone 2016), the portability of wordlists across different discourses within the financial domain is an unresolved issue.

---

[4] The Brown corpus (Francis and Kucera 1979) consisting of 500 samples, each of 2,000 words, from 1960s American written English was one of the first machine-readable text corpora. The British National Corpus (http://www.natcorp.ox.ac.uk/) comprises 100 million words from the 1990s, with a recent update in the BNC2014 project.
[5] Our survey of 149 published articles and working papers reveals just 11 studies (7.4%) that use language from more than one source.

Corpus creation in AF is also limited by the extant focus on U.S. financial discourse. U.S. disclosures are not necessarily representative of international financial language because regulations, business practices and cultural norms vary across countries even when the underlying language (e.g., English) is held constant. Consistent with this view, Brochet et al. (2016) demonstrate how the linguistic properties of non-U.S. firms' English-language conference calls differ from standard business language spoken by native English speakers. There exists a need for AF researchers to study the properties of non-U.S. financial discourse to determine the transferability of popular wordlists and, if necessary, to develop new country-specific versions.

3.2 Corpus annotation

Language description (in linguistics) and NLP (in computer science) typically involve adding information to a raw corpus. The process is known as annotation, labelling or tagging (Garside et al. 1997) and may involve manual or automated approaches. Automated annotation involves labelling elements of text using a predetermined algorithm. A common type of automated annotation is part-of-speech (POS) tagging that classifies each word according to its major word class (e.g. noun, verb, adjective, etc.), along with additional information (e.g., singular or plural, common or proper noun, etc.). POS tags assist with word sense disambiguation and therefore help analyses move from a pure bag-of-words treatment towards a more semantic approach. For example, knowing whether "cost" is being used as a noun or a verb is likely to be critical for interpreting meaning in financial discourse. Accordingly, POS tags are widely used as inputs to disambiguation and feature extraction strategies in CL.

Other types of annotation involve more complex aspects of language such as morphology, grammar, syntax, semantics, pragmatics, and discourse. Table 1 provides details of common taggers in CL. However, whereas POS categories are generally well understood and their corresponding tagsets are highly stable, consensus over the set of appropriate labels declines for higher annotation levels. Automated annotation methods do not feature routinely in mainstream AF work, with Purda and Skillicorn (2015) being a notable exception. Failure to apply this fundamental CL principle casts a shadow over the AF literature, particularly given the dominance of wordlist approaches to feature detection and the attendant disambiguation problems plaguing context-independent dictionaries.

Automated tagging procedures are typically derived from manual annotation. Manual tagging is the starting point in many CL applications for measuring specific features such as sentiment. The approach often follows an iterative process where initial features are refined as knowledge of the phenomenon improves. Once a feature is clearly defined and a robust manual measurement system exists, NLP tools using probabilistic or rule-based methods can be constructed to replicate the annotations automatically (to an acceptable level of accuracy) for a much larger corpus than it is possible to classify manually in any reasonable amount of time (Li 2010a). Examples of AF studies that rely on manual annotation as the starting point for identifying linguistic features in a large sample include Li (2010b) and Huang et al. (2014) for tone, Kravet and Muslu (2013) for risk disclosures, Dikolli et al. (2017) for CEO integrity, and Athanasakou et al. (2019) for strategy-related commentary. However, only the last two studies in this list make their annotations public for other researchers to use and develop.

Manual annotation is inherently interpretative and often subjective. The best results are obtained when multiple human coders annotate the same small set of text independently, compare results, resolve disagreements, and create or update annotation guidelines before continuing to independently annotate a much larger set of texts. Metrics such as Krippendorff's alpha and Cohen's kappa measure agreement between annotators and hence replicability. [See Artstein and Poesio (2008) for a survey of relevant work]. Low levels of agreement indicate the task is poorly defined or the guidelines require refinement (Leech and Smith 2000).

Comparing manual annotation strategies in AF with best practice in CL reveals several concerns. First, we are unaware of any published study where multiple coders annotate the *same* set of text independently to determine replicability. Second, studies do not publish coding guidelines explaining the annotation task. Instead (or at best), studies report selective examples of text to illustrate the classification task in action. This lack of transparency masks the inherently interpretative and subjective nature of the annotation process and creates an illusion of objectivity. The standards applied in studies using CL methods contrasts with papers employing manual content analysis methods that typically use multiple coders, report inter-coder agreement, and provide comprehensive details of coding rules to address concern over subjectivity (e.g., Milne and Adler 1999, Breton and Taffler 2001, Aerts 2005). The lack of transparency associated with CL papers in mainstream AF journals is at odds with the high standards of rigor applied to other aspects of the research design such as identification and raises concerns about replicability and objectivity. (See section 4.4 for further discussion.)

3.3. Natural language processing

Natural language processing (NLP) sits at the core of computational linguistics. Liddy (2001) defines NLP as a suite of computational methods for analysing and representing naturally occurring texts at one or more levels of linguistic analysis to enable human-like processing for various tasks and applications. NLP is considered a subdiscipline of Artificial Intelligence (AI). At the broadest level, NLP methods can be partitioned into machine learning and other AI techniques (Fisher et al. 2016). Critically, NLP involves more sophisticated techniques than simple word frequency counts. Fisher et al. (2016: 162) provide a non-exhaustive overview of NLP methods grouped into three broad categories: supervised systems, unsupervised systems, and semi-supervised systems. Supervised systems require human intervention, which may take various forms including annotating data for a sample of manually-coded positive and negative sentences (e.g., Li 2010b) or selecting corpus features using criteria such as Information Gain (e.g., Goel et al. 2010). The main cost of developing supervised systems is the intervention time required. In contrast, unsupervised systems rely on algorithms that learn how to group unannotated data automatically without requiring human intervention. The process involves grouping similar data points based on pattern matching or clustering (e.g., Balakrishnan et al. 2010, Frankel et al. 2016, Dyer et al. 2017). Finally, semi-supervised systems use a combination of annotated and unannotated data.

Several insights are apparent from Fisher et al.'s (2016) list. First, the suite of NLP technologies is extensive and evolving. Second, NLP techniques used in studies published in mainstream AF journals are restricted to a small subset of available methods. The only NLP
methods currently featuring in multiple papers published in top-tier AF journals are term
weighting (Lourghan and McDonald 2011, Jagadeesh and Wu 2013, Athanasakou et al. 2019),
Naïve Bayes classification (Antweiler & Frank 2004, Li 2010b, Huang et al. 2014), cosine
similarity (Hanley and Hoberg 2010, Brown and Tucker 2011, Hoberg and Lewis 2017), and
topic modelling using Latent Dirichlet Allocation (Campbell et al. 2013, Dyer et al. 2017, Huang
et al. 2018). Instead, the vast majority of mainstream AF research falls into the category of
automated content analysis rather than NLP. Panel A in Figure 1 classifies 149 AF studies
appearing since 2004 by the processing method used.[6] Only 30% of studies summarized in Panel
A employ a material amount of NLP, broadly defined. Basic bag-of-words content analysis
methods dominate, with over 56% of studies relying exclusively on readability or simple
keyword counts. AF researchers must be careful therefore when claiming to use NLP methods
because often such assertions are erroneous. More generally, Panel A of Figure 1 points to a low
level of NLP sophistication in mainstream AF research, with most papers relying on the
econometric equivalent of a two-sample t-test.

      A third insight from the set of NLP methods listed in Fisher et al.'s (2016) review is that
a large body of research applying sophisticated NLP methods to study financial language exists
on the periphery of mainstream AF, most notably in journals such as *Journal of Emerging
Technologies in Accounting*, *International Journal of Accounting Information Systems* and
*Intelligent Systems in Accounting, Finance and Management*. Mainstream AF journals (as defined by tenure lists at international business schools) rarely cite this body of work, suggesting it is off-radar for many researchers. Caution is therefore required where studies claim an incremental contribution through their application of "novel" NLP methods. A substantial body of work applying NLP to financial data also exists in other fields including computer science, operational research, economics, management, and strategy. We therefore conclude that mainstream AF research appears to be behind the curve in terms of its level of NLP sophistication when judged against computer science, other business-related disciplines, and even specialist subfields within AF.

---

[6] The sample comprises 148 accounting and finance papers that apply a significant element of CL analysis, broadly defined. The sample includes published articles and working papers appearing since 2004. Criteria for inclusion in the sample are: (1) the study is published in a mainstream accounting and finance journal (or a journal with a distinct accounting and finance track) appearing on the tenure lists of international business schools; or (2) the study is a working paper available on the Social Science Research Network with a least 10 downloads and at least one co-author from an accounting and finance department in a research intensive business school. The list of 149 papers is available at http://ucrel.lancs.ac.uk/cfie/.

The bulk of NLP work applies supervised text classification, which is the process of learning from previous examples to classify text into certain categories. Examples include named entity recognition (NER) (see section 7.1), sentiment analysis, and the suite of automated taggers summarized in Table 1. Classification tasks such as sentiment analysis involve binary (positive versus negative) or three-way (positive, negative and neutral) outcomes, whereas NER classifies multiple categories (e.g., person names, places and organisation) and POS taggers classify over 100 categories (e.g. nouns, pronouns, verbs, adjectives, articles, determiners, locative, adverbs, etc.). Features extracted using these NLP procedures may serve as inputs to other NLP classification tasks. For example, Purda and Skillicorn (2015) use a POS tagger to identify a feature set that forms the input to their support vector machines (SVM) classifier for detecting fraudulent reports.

Among the group of studies identified as using NLP in Figure 1 (Panel A), the majority use supervised machine learning for sentence- or document-level classification. Examples include: Antweiler and Frank (2004) classifying investment recommendations; Li (2010b) measuring tone in forward-looking MD&A sentences; Huang et al. (2014) classifying tone in analysts' reports; Sprenger et al (2014) measuring sentiment in stock-related twitter messages; and Goel et al. (2010) and Purda and Skillicorn (2015) classifying fraudulent reporting. Naïve Bayes is the default (and sole) classifier used in most AF studies. The argument typically presented in favour of Naïve Bayes is that it is relatively straightforward to apply and displays consistently good classification accuracy even where the conditional independence assumption is violated. Nevertheless, evidence indicates that alternative classifier options such as SVM, Random Forests, and Logistic Regression may outperform Naïve Bayes in particular situations (Domingos and Pazzani 1997, Goel et al. 2010). (See section 6.2.2 for further discussion on text classification.) The narrow focus on a single classification method is indicative of the limited horizons that characterize the use of supervised machine learning techniques in the AF domain more generally. Overall, work in this area lags well behind best practice in the NLP field.

NLP researchers use unsupervised learning methods based on a clustering algorithm such as K-mean in cases where little or no training data exist. The approach scans unannotated data to identify patterns such as co-occurring words that can be used to group text into discrete clusters (with the number of clusters determined by the researcher or the algorithm). One unsupervised learning method gaining traction in AF is topic modelling using Latent Dirichlet Allocation (LDA) (Blei et al. 2003). Example studies include Ball et al. (2015), Dyer et al. (2017) and Huang et al. (2018). The method assumes a generative theory of discourse where content is created by selecting words from a series of 'topic buckets'. Clearly this is not how a discourse is produced and researchers cannot observe topics directly; only the final texts are observable. Topic modelling therefore provides a means of extrapolating backwards from a set of texts to infer the topics that could have generated them. A topic in a topic model is characterized by a list of co-occurring words, with the researcher selecting a suitable label to describe the list. The same word may appear in multiple topics, and in some cases the topics may be more about the genre or style of the discourse than actual content-bearing words that might more usually be viewed as a topic. Underwood (2012) and Murakami et al. (2017) describe the LDA method and provide example topic lists, and Graham et al. (2012) present a tutorial on how to implement LDA in the MALLET software. Linguists are suspicious of LDA for at least three reasons. First, the method relies on a bag-of-words model that does not take account of semantically meaningful multiword expressions or even different meanings of single words. Second, LDA is non-deterministic, meaning that repeating the same process multiple times on the same dataset can generate different topic word lists (Gillings 2016, Hardie 2017). Third, choice of topic label and number of topics involves considerable subjectivity. AF applications of LDA often overlook these features and do not employ more recent innovations in the area (see section 6.2.3).

In conclusion, extant mainstream AF research performs poorly in several respects when judged against NLP norms. In particular: the majority of current work falls within the category of automated content analysis rather than NLP; a significant body of research published in specialist AF journals is not cited; and the fraction of studies applying NLP does not always use the most appropriate methods or discuss implementation issues with sufficient transparency.

3.4 Evaluation

A crucial aspect of the CL endeavour is validating methods and empirical constructs (Resnik and Lin 2013).[7] All components of the CL pipeline from text retrieval to feature detection should be subject to formal evaluation. The task usually involves comparing NLP outcomes against an out-of-sample "gold standard" derived from manual annotation. Metrics used to assess the performance of automatic annotation systems include accuracy (defined as the percentage agreement between automatically assigned labels and the gold standard) and error rate (defined as 100 minus the accuracy score). Where multiple outcomes are possible, performance is typically assessed in terms of precision and recall (Junker et al. 1996, Manning and Schütze 2008). Precision measures the fraction of false positives (Type I errors) and is viewed as a measure of exactness or quality. Recall measures the fraction of false negatives (Type II errors) and reflects a measure of completeness or quantity. A measure of overall accuracy is also often computed using the F score, which is equal to the harmonic mean of precision and recall (Van Rijsbergen 1979).[8]

---

[7] While different evaluation approaches are applicable in different settings, several core principles apply. These general principles were surveyed originally in the Expert Advisory Group on Language Engineering Standards (EAGLES 1996). Intrinsic evaluation involves understanding the performance of a specific processing tool. It can be distinguished from extrinsic evaluation which seeks to determine how well a specific processing tool is performing as part of an entire language technology system (e.g., how much difference does a POS tagger make to a complete speech recognition system that allows a user to talk to their smart phone?)

The quality and transparency of evaluation in the AF literature is variable. Formal assessments of text retrieval accuracy are surprisingly rare. Li (2010b: 1060, footnote 6) reports success rates for MD&A extraction of 95% for 10-Qs and between 85% and 90% for 10-Ks based on a random check of 200 filings, and Campbell et al. (2013) report 98% extraction rates using a random sample of 300 filings. El-Haj et al. (2019) report retrieval and classification accuracy rates around 95% for text extracted from U.K. annual reports published as digital PDF files using a random sample of 586 reports. The majority of studies describe the retrieval process but do not conduct formal evaluations of error rates. Failure to document retrieval accuracy creates the impression that extraction is a straightforward, reliable process equivalent to retrieving financial market data from COMPUSTAT and CRSP when in reality this is often not the case [e.g., Davis and Tama-Sweet (2012, footnote 14)].

---

[8] The F score is derived such that $F_\beta$ measures the effectiveness of retrieval with respect to an individual who attaches $\beta$ times as much importance to precision as recall. The $F_1$ score places equal weight on precision and recall, whereas the $F_2$ ($F_{0.5}$) score weights recall (precision) higher than precision (recall).

Evaluation of text-derived features are more common in the literature although still not ubiquitous. A small proportion of studies follow NLP best practice and conduct manual evaluations (e.g., Bonsall et al. 2017, Dikolli et al. 2017, Athanasakou et al. 2019). At the opposite end of the spectrum, formal evaluations are replaced by ad hoc "sanity checks" (Bozanic et al. 2018, Buehlmaier and Whited 2018). The most common evaluation method in mainstream AF research falls between these two extremes and is based on the following regression framework:

$$LF_{it} = \alpha + \sum_{k=1}^{K} \beta_k Covariate_{kit} + \varepsilon_{it} \qquad (1)$$

where *LF* is the language feature of interest, *Covariate* is a vector of *k* firm- and market-level characteristics predicted by economic theory to correlate with *LF*, and $\beta_k$ is the theoretical mapping from $Covariate_k$ to *LF*. Evaluations test whether $\widehat{\beta_k}$ is statistically different from zero with the expected sign (Li et al. 2013, Dikolli et al. 2017, Henry and Leone 2016, El-Haj et al. 2019). While this indirect method is capable of shedding useful light on the properties of text-derived constructs, model (1) is problematic for two reasons. First, researchers cannot rule out the possibility of correlated omitted variable bias in $\widehat{\beta_k}$. Accordingly, econometric standards normally applied to tests of economic hypotheses do not appear to extend to construct validity tests. Second, it is not possible to specify the true value of $\beta$ in most applications and therefore the approach affords no insights on either the sign or magnitude of measurement error. In contrast, manual evaluations against a gold standard provide precise evidence on noise and bias.

A growing number of AF studies develop machine learning classifiers. A key evaluation principle for machine learning is that text used for training should not be used for evaluation. Creating separate training and holdout samples is the simplest and most rigorous evaluation strategy, with model performance assessed against gold standard classification in the holdout sample. The main limitation on this approach is the cost of constructing the holdout sample.

The K-fold cross validation method is the most common evaluation strategy for machine learning classifiers since it enables all available data to inform the training process. The method involves splitting the dataset into K equal-sized sections. The experiment is then repeated K times and on each iteration, a different section (or fold) is used for testing while the remainder are used for training. A typical partition in each iteration involves using 90% of the data for training and the remaining 10% for testing. The process iterates over multiple folds with random partitioning in each fold. The number of folds usually varies between five and 10. However, recent work raises concern over the precision of the K-fold method and instead proposes using J independent K-fold (J-K-fold) cross validations to evaluate performance (Moss et al. 2018).

In sum, mainstream AF research appears to place less weight on evaluation relative to the norm in CL. Practice varies widely across studies, with comparisons against a gold standard being the exception rather than the norm and attention focusing on accuracy rather than bias. Evaluation strategies also vary significantly within the CL pipeline for the same study. For example, Li (2010b) evaluates the accuracy of his Naïve Bayes classifier measuring the tone of forward-looking MD&A sentences, whereas no evaluations are reported for the more primitive step of retrieving forward-looking performance sentences using a simple wordlist approach (see Li 2010b Appendix B). Finally, direct external validation checks are absent for some of the most popular discourse proxies employed in the AF literature including standard readability algorithms and document length (see Figure 1, Panel A). Indeed, little consensus exists in the academic literature regarding the core properties that define clear, high quality communication. We illustrate this point in Appendix B where we compare domain-expert evaluations of annual report quality against Fog index scores. Findings indicate that manual evaluations of high quality reporting do not display reliably lower readability scores as assumed in extant research. The results are perhaps not surprising since our tests also reveal considerable disagreement across domain experts' quality rankings, reflecting the nebulous nature of the latent quality construct. The analysis indicates the risks of relying on popular text-based metrics without establishing construct validity directly. More generally, we conclude that the high level of rigor normally applied to empirical research in AF does not necessarily extend to the evaluation of CL methods.

**4. Assessment against advantages proposed by Li (2010a)**

Li (2010a) compares manual versus automated content analysis. The advantages of high-quality manual analysis include more precise coding and more granular analysis. However, two structural problems offset these advantages. First, data collection costs are high, leading to small sample sizes that may limit generalizability and statistical power. Second, researcher subjectivity that is inevitably required in the manual coding process can create bias and limit replicability. Li (2010a: 145) argues that automated approaches to analyzing text help to resolve both problems and that, as a result, computer-based analysis of text offers significant incremental advantages over manual coding approaches. Once a scoring algorithm has been developed and validated, automated methods enable researchers to score content for large samples of documents quickly and at low marginal cost. All else equal, larger sample sizes increase statistical power and promote generalizability. Further, because it is possible to apply a scoring algorithm consistently across multiple documents, the approach eliminates conscious and unconscious researcher bias, while also helping to minimize random measurement error associated with inconsistent application of manual coding rules. The approach can also promote replication and extension where details of the scoring algorithm and the process used to implement the analysis are reported transparently. Finally, CL methods used in conjunction with large datasets provide an opportunity to extract features or patterns that even the closest firm- or document-level manual reviews cannot detect.

At first sight the advantages proposed by Li (2010a) are unequivocal and uncontroversial. However, closer inspection highlights the conditional nature of such benefits.

4.1 Reduce data collection costs

Data collection costs are determined by the text preprocessing, classification, and construct validation requirements associated with the specific research question. Consistent with Li's (2010a) claim that automated methods facilitate analysis of very large samples of text at relatively lower cost, extraction costs are indeed low for studies employing documents collected from commercial platforms that offer a reliable application programming interface (API) or from a document repository such as EDGAR whose architecture is designed to support web scraping. Similarly, preprocessing costs are relatively low for documents presented in plain text or HTML format with either a very simple structure (e.g., social media posts or media articles) or a highly standardized reporting template (e.g., 10-K and 10-Q fillings). Scoring costs are also negligible when classifying text using established keyword lists (e.g., Harvard-IV-4, Loughran and McDonald 2011) or measuring readability with proven algorithms (e.g., Fog). Finally, validation costs are relatively low when classification accuracy can be evaluated without the need for a manually constructed gold standard corpus (e.g., Li et al. 2013, Henry and Leone 2016).

However, reliance on CL methods does not guarantee a material reduction in data collection costs because the conditions outlined in the previous paragraph describe only a limited set of basic text applications. Relaxing these constraints can increase implementation costs dramatically, to the extent they may exceed those associated with a small sample manual coding design. Consider the following three examples. First, Li (2010b) develops a Naïve Bayes classifier to measure tone in forward-looking MD&A statements. Training the classifier involves coding 30,000 forward-looking sentences manually, which translates to 250 person-hours based on (an optimistic) estimate of 30 seconds to classify and record the average sentence. Second, El-Haj et al. (2019) develop and validate a tool for retrieving and classifying narrative content from U.K. PDF annual reports. Training the retrieval system involves multiple rounds of manual analysis using up to 1,000 reports in each iteration. They also assess retrieval and classification accuracy using a manually constructed gold standard comprising 11,009 sections from a random sample of 586 reports. Third, Dikolli et al. (2017) construct a measure of CEO integrity by comparing causal language in the letter to shareholders with that presented in the corresponding MD&A (where language is more constrained by corporate lawyers). Their approach involves three significant manual steps: extracting CEO shareholder letters from firms' PDF annual reports because the disclosures are not available as an EDGAR filing or in machine readable text via Compact Disclosure; developing a causal reasoning wordlist; and validating the final wordlist by reading a random sample of 5,000 shareholder letter sentences.

These examples illustrate how CL procedures do not necessarily economize on manual data collection costs. In only the most trivial (and potentially least interesting) applications and those involving little input from domain experts is material manual input minimized or avoided altogether. Typically, the level of manual intervention increases rapidly as the research question and the linguistic features of interest become more sophisticated. Ultimately, significant manual intervention by domain experts is an unavoidable feature of most disclosure research in AF irrespective of the specific methodology employed. Accordingly, choice between manual versus automated coding approaches often simply affects the stage in the research pipeline where manual intervention occurs (along with the trade-off between coding precision and the scalability of the scoring process). Both manual and CL approaches likely involve significant data collection and scoring costs when implemented rigorously, with the nature of the specific research question determining the relative costs and benefits.[9]

Although significant cost savings are unlikely to accrue to researchers who initially develop and validate new CL resources in AF, substantial opportunities for cost savings are possible at the aggregate level if these resources are shared amongst research teams. Resources provided by Loughran and McDonald (2011 and 2016) and El-Haj et al. (2019) illustrate the scope for reducing data-related costs through sharing. Provision of corpora, word lists, training data, and script for harvesting, processing and scoring text as part of an open access approach offers substantial benefits to AF research and beyond. As work in this area matures, we hope that more sharing of text resources will become the norm rather than the exception.

4.2 Increase statistical power via larger sample sizes

Even if CL methods do not guarantee material cost savings, their scalability enables researchers to work with much larger samples which in turn can increase statistical power. However, the positive link between statistical power and sample size assumes constant measurement error. The problem for many CL applications is that measurement error can increase dramatically relative to manual extraction and classification. For example, Dikolli et al. (2017) report error rates of 32% when classifying causal sentences using a wordlist, while Li (2010b) reports error rates up to 15% for extracting the MD&A section from the 10-K filing.[10] Accordingly, gains in the signal-to-noise ratio from applying larger samples may be offset by greater measurement error due to noisier extraction and classification. Indeed, it is perfectly possible that more precise manual coding applied to a few hundred observations may generate more power than automated methods applied to tens of thousands of documents as the complexity of the linguistic features or the reliance on naïve bag-of-words methods increases. The idea that automated analysis of very large samples of text guarantees greater statistical power is, therefore, a fallacy. Further, material measurement error also increases the risk of bias when the source of the error is not clearly understood. The allure of large sample sizes associated with automated content analysis, therefore, represents a potentially dangerous trap, particularly for less experienced researchers.

---

[9] We focus on extraction and processing costs from a user perspective. Lewis and Young (2019) discuss how this is (or at least should be) an increasingly important consideration for regulators responsible for developing disclosure rules. Unless regulators acknowledge the increasing importance of automated processing and take steps to reduce retrieval costs, many of the benefits associated with higher quality disclosure may not be realized fully.

4.3. Promote generalizability

The scalability of CL methods helps promote generalizability through analysis of larger and more representative samples. While superficially this claim seems uncontentious, text retrieval and processing restrictions mean that even very large samples may suffer material selection bias in various forms. For example, the desire to minimize extraction costs can induce observational bias in the form of the "streetlight effect" where individuals limit their search activity to locations where it is easiest to look (Kaplan 1964). Figure 1, Panel B, shows how extant AF research focuses on corpora where retrieval costs are low or moderate, such as 10-Ks, earnings announcements and conference calls. While these sources represent a natural focus for AF researchers, the disproportionate concentration on 10-K and 10-Q filings in general, and MD&As in particular, is at odds with concern over the relevance and timeliness of annual reports for economic decision making (Chen and Li 2015). The pattern provides a *prima facie* case that other interesting and economically important narratives may be being overlooked due to higher retrieval and processing costs.

---

[10] Davis and Tama-Sweet (2012: footnote 14) note that 10-K and 10-Q filings extracted directly from EDGAR include a large number of extraneous characters due to different text-file formats used by registrants and EDGAR. Davis and Tama-Sweet use the 10-K Wizard tool to clean raw text files prior to analysis. Their approach contrasts with other studies that use raw text retrieved from EGDAR. Extraction errors are especially common when working with PDF files. For example, El-Haj et al. (2019) discuss how certain character strings in PDF file are systematically corrupted in the retrieval process. Meanwhile, content presented in column format may be scrambled if the extraction process incorrectly reads directly across the page rather than column by column. Such extraction errors can create severe measurement error for sentence-level analysis (including readability metrics).

26Electronic copy available at: https://ssrn.com/abstract=3330757

Generalizability also depends on how the initial population is defined. Where the population is a specific reporting regime then a larger sample enhances generalizability all else equal. However, where the population is defined to include multiple reporting regimes then simply increasing sample size from a single regime does not necessarily yield more generalizable insights. Extant CL work on annual reports provides three examples. First, research using 10-Ks and 10-Qs typically omits firm-years where the MD&A is incorporated by reference to the shareholders' report. As a result, even very large samples of 10-Ks may not be fully representative of the population of filers where firms whose MD&A is incorporated by reference differ systematically on one or more dimensions. Bafundi et al. (2018) provide evidence on the sampling biases introduced when retrieving and processing 10-Ks from EDGAR. Second, the ability to harvest and process large volumes of 10-K filings undoubtedly enhances generalizability *within a U.S. context*. However, since the 10-K reporting template produces annual reports that differ in structure and content from other jurisdictions, the applicability of 10-K evidence to reports published in non-U.S. settings is unclear.[11] Finally, the high concentration of studies using U.S. data risks biasing conclusions if narrative disclosures are shaped by preparers' incentives and institutional features to the same extent as documented for financial statement outcomes (Hail et al. 2010).[12] Similarly, language structure also varies across countries. Kim et al. (2017) predict and find that earnings management activity is less prevalent in countries associated with weaker future-time reference languages.

The procedure used to retrieve and process text may also influence generalizability. For example, Lang and Stice-Lawrence (2015) show that higher quality annual report disclosures, as proxied by IFRS adoption, are associated with positive stock market outcomes. However, their analysis does not distinguish between management commentary and financial statement footnotes because their method for extracting text from PDF annual reports does not capture the location of commentary within each document. El-Haj et al. (2019) analyse U.K. annual reports and distinguish management commentary from the financial statements. Results indicate that increases in narrative content following IFRS adoption are confined to the financial statements component of the annual report and that as a result Lang and Stice-Lawrence's (2015) conclusion is unlikely to generalize to management commentary.

4.4 Improve objectivity and replicability

Manual content analysis inevitably involves researcher subjectivity, which in turn raises questions about replicability. Studies applying manual coding methods typically address this challenge by providing extensive details of the scoring instrument together with examples that demonstrate how the coding procedure is applied. While these steps provide clarity and help to support replication, significant judgement may still be necessary even when implementing the most transparent coding procedure. CL methods are attractive because algorithms apply a set of rules consistently and dispassionately. Results for a single document or an entire sample are, therefore, replicable multiple times and by different individuals.

---

[11] Since the population of non-U.S. annual reports far exceeds the number of 10-Ks, it remains an open question whether extant results are representative of the median annual report globally.
[12] There is good reason to believe these differences may be *more* pronounced for the narrative component of annual reports and earnings announcements because unlike financial statement outcomes, textual disclosures are not bound by the temporal discipline of double-entry bookkeeping. Accordingly, insights emerging from the extant literature are likely to be partial at best.

While the promise of high replicability is an attractive feature of CL approaches, closer inspection highlights the need for caution. First, CL is not immune to researcher subjectivity as discussed in section 3.3. Varying degrees of judgement are necessary for all but the most basic bag-of-words methods. Examples of researcher subjectivity when applying CL methods include: text pre-processing choices such as stemming and removal of stop words (Loughran and McDonald 2016); construction of domain-specific wordlists (Loughran and McDonald 2011, Li et al. 2013, Dikolli et al. 2017, Athanasakou et al. 2019); manual coding of training data in machine learning applications (Li 2010b, Huang et al. 2014); feature selection using Information Gain (Goel et al. 2010); and topic identification in LDA models (Ball et al. 2015, Dyer et al. 2017). Second, replicability is a function of the precision with which various steps and decision rules in the process are communicated to other researchers. Although many AF studies provide transparent information regarding text preprocessing steps (e.g., Li 2008, Grüning 2011, Campbell et al. 2013, Bodnaruk et al. 2015), a fraction of papers fails to provide the detail required to support replication and interpretation of results.[13] Similarly, important

---

[13] Common refinements to the raw corpus include removing proper nouns and stop words, adjusting hyphenated words, correcting for spelling differences between American English and British English, and converting all words to lower case so that otherwise identical words are not included in the corpus as distinct elements. The type of adjustment(s) is conditional in part on the research question and CL method employed, and therefore care is required when making choices. For example, exclusion of stop words is standard practice in machine learning applications to reduce dimensionality of the feature space and improve processing efficiency (e.g., Goel et al. 2010). However, stop

implementation details when using Naïve Bayes classification are not always provided. For example, two distinct but interconnected variations of Naïve Bayes are the multinomial model and multivariate Bernoulli model (McCallum and Nigam 1998). Ren et al. (2013) report differences in classification accuracy for the two models conditional on the size of the feature set, with the multinomial model over-fitting (performing better) when the feature set is small (large). Finally, and in contrast to the CL literature, publication of data and code is not common practice in mainstream AF research.

The need for significant researcher judgement combined with opaque reporting practices mean that the proposed benefits of objectivity and replicability may be illusionary. We view this as a potentially more dangerous situation compared with manual coding designs where subjectivity is acknowledged explicitly. In the latter case appropriate warning signs serve to stimulate healthy scepticism whereas the strong claims of objectivity often made by AF studies using CL methods risk creating a false sense of precision.[14]

## 5. Assessment against practical relevance

Placing extant research within the context of professional and regulatory debates about the provision and use of unstructured data helps shed light on the issue of relevance. We focus

---

words such as personal pronouns may play a central role in some analyses (Larcker and Zakolyukina 2012, Purda and Skillicorn 2015). Similarly, purging the corpus of proper nouns such as company names, products, countries, individuals, months of the year, etc. is standard practice to avoid misclassification of words. However, as many proper nouns are industry-specific, error levels are partly conditional on sample composition. For example, "crude" and "heavy" describe specific grades of oil in the oil and gas sector; and "trust" and "mutual" describe particular organisational types in the financial sector.

[14] Recent work on topic modelling using LDA serves as an example. Ball et al. (2015: 1) state that "The linguistic methods are automated, take very little input from the researcher, and are thus not biased by researcher prejudice" and "the linguistic methods used in our study are fully replicable"; Dyer et al. (2017: 229) state "Because LDA is an unsupervised method, it is replicable and free of researcher bias."; and Huang et al. (2018: 2835) claim that topic modelling "provides a reliable and replicable classification of topics. Neither of these features can be attained with manual coding, which relies on human coders' subjective judgment". However, as the discussion in section 3.3 highlights, LDA involves significant researcher judgement in terms of topic interpretation and does not yield replicable results due to the non-deterministic nature of the model.

primarily on financial reporting and securities market regulation where CL methods offer promising opportunities for AF researchers to contribute to practice.

Financial regulators and accounting standard setters stress the importance of clear, concise, balanced and forward-looking narrative reporting (e.g., SEC 1998, International Accounting Standards Board [IASB] 2017). Applying CL methods to financial disclosures helps mainstream empirical research move beyond viewing disclosure as a simple binary outcome to address more subtle questions about *how* and *where* information is communicated, and how such outcomes impact economic decisions. Work examining language features such as the length, readability and tone reveals important statistical and economic associations that speak to the usefulness of narrative disclosures (Li 2010a, Merkley 2014, De Franco et al. 2014, De Franco et al. 2015, Bonsall et al. 2017, Hsieh et al. 2015). Investor protection is an area where such work is informing practice. Compelling evidence exists that variables derived from unstructured data provide incremental predictive power for fraud and GAAP violations beyond quantitative data (Larcker and Zokolyukina 2012, Purda and Sillcorn 2015, Brown et al. 2017, Hoberg and Lewis 2017). Reflecting these results, the Deputy Director and Chief Economist at the SEC's Division of Economic and Risk Analysis highlighted the scale of the contributions CL is making to its enforcement actions (Bauguess 2016). The SEC has invested heavily in machine learning and NLP technology to measure topic and tonality signals in firms' filings, which are then mapped into established risk factors. Away from fraud detection, the SEC has also used CL methods to assess how emerging growth companies avail themselves of JOBS Act provisions. The scalability benefits provided by CL mean that work in this area is well placed to contribute further to securities market monitoring.

While extant CL results speak to practical financial market issues, several factors mean they may also face a credibility problem with many practitioners. First, the majority of CL research operates at an aggregate level (e.g., the entire earnings announcement; the entire 10-K; the complete MD&A), whereas practitioners are often interested in more granular issues such as the format and content of specific disclosures such as a footnote, the placement of particular content within the overall reporting package (e.g., in the financial statements versus in the narrative section of the annual report versus on the company's website), and limits on the use of jargon, acronyms and industry-specific terms without clear definitions (Financial Reporting Council [FRC] 2015, IASB 2017). Second, common empirical proxies for key reporting features such as understandability, balance and boilerplating are not necessarily recognizable to financial market professionals. For example, regulators may struggle to make the connection between their perception of high quality narrative disclosure and readability indices such as Fog and Bog (Bonsall et al. 2017) that focus on simple language properties such as word and sentence length rather than fundamental features such as the relevance of information to the audience and the way content is organized. Take the properties of ineffective communication identified in IASB's Discussion Paper on the principles of disclosure (IASB 2017: 19-20, para 2.4).[15] Precisely how such features map into CL constructs such as Fog is unclear, particularly given the lack of direct evaluation evidence (see section 3.4).

The challenge for CL applications to be viewed as practically relevant is further illustrated by the following two financial reporting examples. First, the IASB's practice statement on management commentary highlights two underlying reporting principles: providing management's view of the entity's performance, position and progress; and supplementing and complementing information presented in the financial statements (IASB 2010: para 12). In the IASB's view, ensuring alignment with these principles requires management commentary to include information that meets the qualitative characteristics described in the IASB's conceptual framework (IASB 2018). While properties such as relevance and timeliness may be amenable to investigation using CL methods, characteristics such as comparability, understandability and verifiability are more challenging to operationalize automatically.

---

[15] According to the IASB (2017), ineffective disclosure involves: generic or boilerplate descriptions; use of unclear descriptions and undefined technical jargon; poor organisation of information including failure to provide a contents page or other navigation aids; unclear linkage between related but diffuse pieces of information; unnecessary duplication of information; disclosing information in a format that is inconsistent with industry practice or changing the way information is disclosed from period to period; using narrative disclosure when a table would be more effective; and omitting material information or including immaterial information that may obscure material content.

Second, the view that high quality financial reporting should involve a clear narrative that integrates various elements of business performance and practice around issues of sustainability and long-term value creation is an emerging theme among regulators (IASB 2010, International Integrated Reporting Council 2011, European Financial Reporting Advisory Group 2013, FRC 2014). Studies using CL methods are yet to tackle issues of organisation and integration. While approaches such as measuring the degree of cross-referencing may contribute to this debate, the scope for developing large sample proxies for features such as the level of integrated commentary and the quality of business model reporting is limited in our view.

Overall, the scope for CL methods to outperform manual inspection for nuanced aspects of disclosure such as materiality, integration, comparability, understandability and clarity is questionable since these concepts prove challenging for humans to define and measure consistently. The justification for viewing CL methods as the default approach for studying policy-relevant aspects of unstructured data, therefore, appears weak, with (less fashionable) manual analysis using smaller samples affording significant advantages in many cases. It is important that AF researchers are sensitive to how the allure of large datasets and the low entry costs associated with text constructs such as readability and tone may serve to drive a further wedge between research agendas and practical relevance.

## 6. Common themes, innovations and incremental contribution

This section summarizes the current state of play in mainstream AF research with respect to the use of CL methods. Our goal is not to provide a comprehensive review of extant work. Instead, we aim to summarize broad trends relating to the themes examined and core methods employed, critique more innovative applications of automated discourse analysis, and assess incremental contribution relative to manual analysis of financial discourse.

6.1 The prevailing landscape number

Panels A-C in Figure 1 highlight the dominant trends characterizing CL approaches in mainstream AF research. The typical study examines 10-K filings (Panel B) using basic automated bag-of-words content analysis methods such as readability algorithms and keyword counts (Panel A). For the 56% of the sample applying simple word count technology, the majority of papers (65%) measure tone, with risk and uncertainty a distant second (Panel C). This level of clustering is surprising given the variety of financial discourses, CL methods and unresolved questions regarding the properties and role(s) of unstructured financial data, coupled with the importance of incremental contribution as a determinant of publication success in AF journals. The pattern is consistent with the initial phase of the research lifecycle. A concern with this first phase, therefore, is the degree to which agendas are shaped by ease of data access and empirical implementation rather than research priorities.

Loughran and McDonald (2016) review the literature on readability and automated keyword content analysis. A key theme in their analysis concerns the benefits of parsimony. Simple bag-of-words methods provide an efficient solution relative to more complex NLP methods when the sequence of words in a discourse or the sense in which words are used are not critically important to measuring the attribute of interest accurately. Consistent with the view that significant progress is possible using fairly primitive content analysis methods, Henry and Leone (2016) compare alternative approaches to measuring tone in earnings-related announcements and find that parsimonious domain-specific word lists such as Henry (2006, 2008) perform well relative to more comprehensive wordlists such as Loughran and McDonald (2011) and more sophisticated CL methods such as machine learning classification using Naïve Bayes. Parsimony and simplicity also promote transparency and replicability.

The main problem with simple constructs such as readability and keyword measures of tone is that word sense and sequence *are* critically important for many (arguably most) aspects of financial discourse. In the same way that stock returns are difficult to interpret without adjusting for risk and revenues do not provide reliable performance insights until expenses and capital investment are considered, so word sense and sequence are essential for interpreting meaning in financial discourse. In our view, acknowledging this fact is critical to ensuring that the second phase of CL work in AF delivers robust and relevant findings that inform theory and practice. As illustrated further in the critique of topic modelling presented in section 6.2.3, capturing meaning in financial discourse is more challenging than simply adopting more sophisticated NLP technology because many such approaches still rely on a bag-of-words structure. The same holds when isolating the abnormal component of constructs such as tone (Huang et al. 2014) and readability (Bushee et al. 2018): decomposing bag-of-words measures into normal and abnormal components does not correct for the failure to consider word sense in the underlying construct.

Building on the insights provided by the first wave of CL-related work on financial discourse therefore requires mainstream AF research to evolve (quickly) from the prevailing norms identified in Figure 1 to embrace analysis of word sense and meaning. Promoting research agendas where the focus or approach continues to rely on bag-of-words technologies such as readability and basic keyword content analysis, will at best represent a missed opportunity and at worst yield a body of work whose credibility and relevance is questioned by future generations.

6.2 Moving beyond basic content analysis

6.2.1 Refined unigram word count applications

The predisposition toward simple (and often domain independent) content analysis methods evidenced in Panel A of Figure 1 conceals refinements and promising avenues pursued in a significant fraction of studies. The approaches used in this subset of studies reveal how even basic word-level methods can be used to provide richer insights into the properties and effects of financial discourse. The following discussion reviews four examples.

Our first example involves two innovative refinements to standard readability scores such as the Fog and Flesch indices. The first refinement is the Bog index derived from the *StyleWriter* software package (Bonsall et al. 2017). Bog captures the plain English attributes of disclosure highlighted by linguists (e.g., active voice, fewer hidden verbs, fewer abstract words, etc.) and is therefore more closely aligned with notions of clear and concise discourse promoted by regulators (SEC 1998, FRC 2015). Further, rather than treating all words of three or more syllables as complex as is the case with Fog, complexity is determined by word familiarity using a proprietary list of 200,000 words representative of natural language. Bog therefore helps to address the problem of determining complexity based on syllable counts alone (Loughran and McDonald 2014a). Bonsall et al. (2017) provide robust evidence for manual evaluations and statistical correlations that Bog outperforms Fog, Loughran and McDonald's (2014a) plain English index, and more primitive proxies such as document length and file size.

The second readability refinement is Bushee et al.'s (2018) decomposition of Fog into an information component reflecting the provision of complex (but useful) information and an obfuscation component reflecting managerial opportunism. The information component is the fitted value from regressing Fog scores derived from management conference call discourse on Fog scores derived from analysts' language during the same call. The unexplained component of management Fog, as reflected in the regression error, is then interpreted as the obfuscation component. Validations based on correlation tests suggest that the approach is successful at distinguishing inherently complex content from obfuscation.

Although the methods proposed by Bonsall et al. (2017) and Bushee et al. (2018) still rely on a bag-of-words view of discourse, both approaches embed aspects of corpus linguistic analysis and as such represent welcome departures from naive unigram word counts. For example, Bog focuses on specific attributes of discourse that have been shown by linguists to influence clarify and understandability, while both studies measure distinct discourse properties using a reference corpus. Results illustrate how thoughtful yet relatively straightforward adjustments based on linguist theory can yield material improvements in even the simplest methods of quantifying discourse.

Our second illustrative approach relates to the use of word-level weighting procedures. Normalizing raw word counts to avoid a mechanical correlation with discourse length, improve disambiguation, and increase classification performance (e.g., giving more weight to unusual words) is standard practice in many NLP applications (e.g., Balakrishnan et al. 2010). Loughran and McDonald (2016: 1207-1209) provide an overview of the topic. Despite the intuition for using weighting procedures, most AF studies scale word counts by document length to mitigate size effects but ignore other benefits of weighting such as enhanced word sense disambiguation. Examples of this more refined approach to implementing bag-of-words methods nevertheless exist. Loughran and McDonald (2011) and Brown and Tucker (2011) use the term frequency-inverse document frequency (*tf.idf*) method. Meanwhile, Jadadeesh and Wu (2013) and Athanasakou et al. (2019) develop new weighting approaches. Specifically, Jagadeesh and Wu (2013) define weights (and polarity) as coefficients from regressions of market reactions on positive and negative words from firms' 10-K fillings, while Athanasakou et al. (2019) use a corpus method to derive conditional probabilities that words in their strategy dictionary are predictive of observable strategy-related discourse. Both Loughran and McDonald (2011) and Jagadeesh and Wu (2013) provide evidence that weighting improves performance relative to scaling by total word count. Indeed, Jagadeesh and Wu (2013: 712) conclude that "appropriate choice of term weighting in content analysis is at least as important as, and perhaps more important than, a complete and accurate compilation of the word list". In contrast, Henry and Leone (2016) find no evidence that *tf.idf* weighting improves performance of their tone metrics.

Domain-specific wordlist construction is our third area where extant work reveals interesting applications of basic keyword content analysis. Research provides strong evidence that domain-specific wordlists outperform general language dictionaries (Loughran and McDonald 2011, Henry and Leone 2016). Henry (2008) and Loughran and McDonald (2011) use corpus methods to create domain-specific wordlists. For example, Loughran and McDonald (2011) create a series of domain-specific wordlists from a corpus of 10-Ks filed during 1994 to 2008. All words occurring in at least 5% of documents in the corpus are examined manually to determine their most likely usage in financial discourse and resolve lexical ambiguity. Campbell et al. (2013) also use corpus methods to construct their list of risk factor keywords from 10-K fillings. Their approach involves two steps. First, an initial domain-specific list of risk-related words is identified from prior research (Nelson and Pritchard 2007). This list is then augmented in step two using LDA topic modeling to identify additional words from their 10-K corpus that consistently appear in firms' risk factor sections.[16] In contrast to these bottom-up corpus approaches, other studies use a top-down method to library construction whereby candidate *n*-grams are identified from domain- and topic-specific resources such as textbooks (e.g., Hanley and Hoberg 2010, Hassan et al. 2018, Athanasakou et al. 2019). No study of which we are aware uses a combination of bottom-up and top-down methods to isolate the subset of common words.

Our final example of an interesting application of basic keyword content analysis relates to discourse filtering. Most wordlist applications count keywords appearing in an entire discourse (e.g., 10-K or MD&A). Using this unfiltered approach increases the risk of Type I errors because key words may appear in contexts unrelated to the construct of interest. For example, while "below" features regularly in negative keyword lists (e.g., Henry 2006, 2008), it may also be used to refer to the location of other content in a report. A simple disambiguation technique involves using keywords to isolate sentences in a discourse that relate directly to a construct of interest. For example, Li (2010b) and Anthanasakou and Hussainey (2014) use a

---

[16] A weakness of Campbell et al.'s (2013) novel approach from a corpus linguistics perspective is that their method of identifying words that consistently appear in firms' risk factor sections does not benchmark against language in a reference corpus of 10-K sections that do not contain material disclosures on risk factors. As a result, it is not clear whether the additional words identified by Campbell et al. (2013) have discriminatory power for risk-related content. See section 7.4 for details of corpus methods such as keyness that identify the distinguishing features of a discourse.

keyword approach to identify forward-looking statements in 10-K and U.K. annual reports, respectively. Identified sentences are then scored on one or more dimensions (e.g., tone) to yield more fine-grained measures of discourse properties. In addition to improving classification performance, filtering can lower manual classification costs significantly by reducing the volume of content for human coders to review. For example, Bentley et al. (2018) and Black et al. (2018) use CL methods to extract sentences referencing non-GAAP performance measures, which are then analyzed manually in a second stage. The approach therefore highlights how automated and manual methods can serve as complements in financial discourse analysis. Critical to the success of this filtering method is the accuracy with which target sentences are identified (recall) and irrelevant sentences are excluded (precision). As discussed in section 3.4, formal evaluations of precision and recall for such applications are rare in mainstream AF research.

6.2.2 Machine learning classifiers

Evidence from the NLP literature consistently demonstrates that supervised machine learning methods can produce more accurate classification of discourse properties than simple content analysis approaches such as counting keywords (Ng and Zelle 1997). Machine learning text classifiers have been used sparingly in AF to measure tone (Antweiler and Frank 2004, Das and Chen 2007, Li 2010b, Huang et al. 2014, Sprenger et al. 2014, Bartov et al. 2018), deception and fraud (Goel et al. 2010, Purda and Skillicorn 2015, Hoberg and Lewis 2017), and financing constrains (Buehimaier and Whited 2018). With rare exceptions (e.g., Goel et al. 2010, Purda and Skillicorn 2015, Gao and Lin 2015), most AF applications rely on a Naïve Bayes model. Although classification performance with Naïve Bayes is encouraging, theory provides little guidance as to which specific classification method is likely to perform best in a given setting.

Typically, therefore, NLP researchers compare a range of methods to determine the best performer. For example, Goel et al. (2010) find that SVM displays superior classification accuracy in their fraud detection analysis, with Naïve Bayes failing to beat the random baseline. Exclusive reliance on a signal classifier in most AF studies contrasts with the battery of econometric robustness tests reported by empirical studies in the field.

Significant scope also exists to refine the way classification algorithms are implemented. AF applications often rely on the most basic implementation procedure, which limits classification performance and fuels doubt over the net benefits of using machine learning given the incremental costs relative to naïve keyword content analysis methods (Henry and Leone 2016). The following three examples highlight the opportunities to improve implementation and enhance performance. First, AbuZeina and Al-Anzi (2017) note that high dimensionality leading to sparse data is a common problem for algorithms that reduce discourse to a vector of word counts (Goel et al. 2010, Li 2010b, Buehimaier and Whited 2018). While methods such Information Gain are used frequently in NLP to help reduce dimensionality (Goel et al. 2010), they rarely feature in extant mainstream AF applications. Second, the choice between using a balanced versus unbalanced training sample can have a dramatic impact on out-of-sample classification accuracy (Bhowan et al 2013).[17] The issue is rarely discussed in mainstream AF studies, suggesting that all training data are used as part of an unbalanced approach.

---

[17] A balanced approach to specifying the training sample ensures equal sample sizes for each category being classified. For example, a balanced approach to training a sentence-level tone classifier will ensure the same number of positive versus negative sentences. In contrast, an unbalanced approach uses all available data and therefore risks biasing classification towards the majority case in samples where one category is empirically dominant (e.g., positive sentences in management performance commentary). Under-sampling (i.e., balanced sample approach) can fail when observations on a class are particularly sparse. The synthetic minority over-sampling technique (SMOTE) provides a means of dealing with sparse data (Chawla et al. 2002).

A third aspect where the typical AF approach tends to be restrictive relates to the selection of text features as inputs to the classifier algorithm. While a vast array of features serves as candidate inputs, the majority of AF applications rely on a limited range of surface-level features such as word counts and sentence complexity (Goel et al. 2010, Buehimaier and Whited 2018). Machine learning classifiers applied in AF research therefore often face the same fundamental weakness as simple content analysis using keywords insofar as they treat discourse as bag-of-words that ignores context and meaning. Expanding the scope of the feature set to include more advanced syntactic and semantic features (see sections 3.2 and 7.4), along with domain-specific wordlists (e.g., Henry 2008 Loughran and McDonald 2011) and weighting methods such as *tf-idf* can improve disambiguation and increase classification performance (Iyyer et al. 2015, Bloehdorn and Moschitti 2007, Loughran and McDonald 2016).

6.2.3 Topic modelling

Section 3.3 highlights a trend toward topic modelling in mainstream AF. Using NLP methods to detect patterns in unstructured data that can be linked to specific economic themes opens the door to a range of exciting opportunities as illustrated by the pioneering work of Campbell et al. (2013), Dyer et al. (2017), Bozanic et al. (2018) and Huang et al. (2018). Work is nevertheless characterized by the same narrow methodological focus as documented for text classification insofar as LDA is presented as the sole procedure for extracting topic categories. Section 3.3 summarizes the well-known weaknesses associated with the LDA method, notably the non-deterministic nature of the model that renders replication extremely difficult despite claims to the contrary (Chuang et al. 2015).

New topic modelling procedures have been developed in the NLP literature to address some of the problems associated with LDA, particularly in relation to non-determinism. Viable (and in some cases superior) alternatives to LDA include Correlated Topic Modelling (Blei and Lafferty 2006), Probabilistic Latent Semantic Analysis (Hofmann 1999), and Structural Topic Modelling (Roberts et al. 2013). Note, however, that many fundamental criticisms of LDA beyond non-determinism also apply to these approaches, such as ignoring both macro-level document structure and micro-level constituent structure (Baldwin 2011), and degradation of topic stability due to stemming (Schofield and Mimno 2015). Further, Aletras and Stevenson (2014) report that metrics for measuring topic similarity may be sensitive to high dimensionality, the number of topics, and size of the corpus. Collectively, these concerns suggest that AF researchers must proceed with caution when applying topic modelling procedures.

Finally, and arguably most critically, existing topic models are based on a bag-of-words (i.e. unigram word list) structure that does account for semantically meaningful multiword expressions or different meanings of the same word. Linguists are therefore justifiably suspicious of such models. Murakami et al. (2017) suggest using the key semantic domains method as an alternative approach to standard NLP methods (Rayson 2008), as well as Multi-Dimensional Analysis (Biber 1988). The message for AF researchers with respect to topic modelling is therefore the same as other CL applications: validity is ultimately determined by how well or otherwise the method is able to capture meaning effectively.

6.3 Incremental contribution relative to manual analysis

There is little doubt that the use of CL methods represents a game-changing development for mainstream AF researchers. Nevertheless, care is required to avoid overstating its contribution while understating the value of alternative methodologies. Research on financial discourse did not begin with Tetlock (2007), Li (2008), Loughran and McDonald (2011) and similar early adopters of CL technology. Acknowledging this fact is critical to evaluating the contribution of studies applying CL methods.

Research examining the properties, determinants and economic consequences of financial discourse has a long tradition in AF. Jones and Shoemaker (1994) and Merkl-Davies and Brennan (2007) review work in the area, the bulk of which relied on manual coding applied to samples comprising hundreds (rather than thousands) of observations. Work focuses primarily on understanding the properties of financial narratives and testing how reporters' incentives shape disclosure outcomes. In contrast to recent CL-focused work, a high fraction of these studies examines non-U.S. discourse. Two contrasting hypotheses underpin much of the work reviewed by Merkl-Davies and Brennan (2007). The information view predicts that reporters use discourse to communicate information that complements and supplements quantitative disclosures. The alternative perspective predicts that narratives display strong self-serving reporter bias due to either conscious efforts to manage receivers' impressions or as a result of unintentional cognitive bias.[18] Studies using manual content analysis provide compelling evidence in support of both views, with outcomes in a given setting conditioned by the strength and direction of reporters' incentives (Clatworthy and Jones 2003, Aerts 2005, Baginski et al. 2004, Schleicher and Walker 2010, Schleicher 2012, Keusch et al 2012, Kimborough and Wang 2014). The volume of evidence, coupled with fine-grained manual coding schemes that facilitate analysis of language at the syntactic, semantic, pragmatic, and discourse levels ensures that this body of work cannot be dismissed on the grounds of researcher subjectivity, lack of generalizability, or low statistical power. Indeed, powerful arguments exist for viewing manual analysis as the default approach to studying the properties of financial text given the complex nature of language and communication. In light of this extensive body of rigorous evidence, the process of establishing incremental contribution for studies adopting a CL lens to studying financial discourse is not a straightforward task.

---

[18] European disclosure researchers have traditionally referred to such behaviour as "impression management" whereas U.S. researchers tend to prefer terms such as "strategic reporting" and "obfuscation".

Many CL-focused studies re-examine broadly similar issues to those reviewed by Merkl-Davies and Brennan (2007) and arrive at broadly similar conclusions. Evidence from large sample CL studies supports both the information role of financial discourse (Li 2010b, Feldman et al. 2010, Brown and Tucker 2011, Davies et al. 2012, Merkley 2014) and the managerial opportunism or obfuscation perspective (Li 2008, Cho et al. 2010, Huang et al. 2014, Davis and Tama-Sweet 2012, Allee and DeAngelis 2015). Other than relying on of a different research design, incremental contribution beyond the insights provided by extant manual analysis is not always apparent in this body of work given the similarity in research questions. Greater care is therefore required in emerging CL work to ensure that evidence from extant manual content analysis studies (often based on non-U.S. data) is not overlooked or dismissed as unreliable when establishing incremental contribution. This is particularly important in light of the inevitable incremental costs resulting from increased measurement error.

Not all discourse-related research questions and settings in AF lend themselves naturally to the CL treatment. For example, many important research questions involve disclosures that are hard to process automatically due to the complexity of the content, the text source preventing reliable retrieval of all relevant information, or a combination of both. Examples include: the use of unstandardized alternative performance measures in the annual report (Voulgaris et al. 2014,

European Securities Markets Authority 2015, Guillamon-Saorin et al. 2017); the interaction between text, graphs and pictorial content (Davidson 2008, Beattie and Jones 2008); the role of management commentary explaining unusual changes in financial statement line items such inventory (Sun 2012); the fine-grained features of analysts' reports (Asquith et al. 2005); and the plausibility of seemingly self-serving attributions in earnings press releases (Kimbrough and Wang 2014).[19] CL methods by construction are also likely to struggle in settings where the focus is on nondisclosure or silence. For example, Hollander et al. (2010) use manual methods to examine the causes and consequences of management choosing not to answer participants' questions during earnings conference calls.

Despite these caveats, recent work employing CL methods provides numerous new and important insights that extend beyond the scope of manual analysis. The most significant contributions involve research questions that align closely with the fundamental benefits of CL in the form of scalability and latent feature detection. For example, consequence studies that examine the average effect of financial discourse and cross-sectional variation therein typically require larger sample sizes to generate sufficient statistical power, relative to incentive studies that examine variation in discourse properties. Adoption of scalable CL methods therefore enables researchers to develop sharper insights regarding the economic consequences of financial discourse (e.g., Brown and Tucker 2011, Lehavy et al. 2011, Lawrence 2013, Brochet et al. 2016, Bonsall and Miller 2017). Scalability coupled with latent feature detection is also helping researchers develop new empirical proxies for constructs such as obfuscation (Bushee et al. 2018), CEO integrity (Dikolli et al. 2017), behavioral bias (Mokoaleli-Mokoteli et al. 2009), and geographic dispersion (Platikanova and Mattei 2016) that have proved challenging or impossible to measure with quantitative data alone. Collectively, these contributions are raising the profile of unstructured data in AF research to match the status traditionally afforded to quantitative outputs. This shift in focus is essential since the majority of new data involves a high fraction of unstructured content.

---

[19] Several papers examine causal language and attributions using CL methods (Zhang and Aerts 2015, Zhang et al. (2019), Koo et al. 2017, Dikolli et al. 2017). While these studies provide useful insights on aspects of causal reasoning, they focus primarily on measuring the incidence and volume of causal language using causation keywords rather than on distinguishing between defensive and enhancing attributions or identifying the nature of specific causal factor(s).

We view the large-sample CL approach as offering important new insights and opportunities beyond traditional manual approaches to scoring financial discourse. Critically, however, while CL appears to be replacing manual scoring methods as the default research design for financial discourse analysis, neither the insights afforded by extant manual analysis nor the method itself are rendered redundant. Manual analysis still has a central role to play in the study of unstructured financial data. We therefore stress the importance of viewing CL and manual methods as complementary approaches in the research toolkit. Bentley et al. (2018) and Black et al. (2018) are examples of this complementarity. For example, Bentley et al. (2018) use CL methods to extract non-GAAP sentences from firms' earnings press releases and then use manual analysis to retrieve non-GAAP earnings information from selected sentences. Accordingly, the key challenge for those choosing the CL route is to ensure that the research question aligns closely with the comparative advantages of the approach, rather than relying on vague and unconvincing references to enhanced generalizability, power and replicability.

**7. New horizons in textual analysis**

This section highlights four tools from the CL literature that assist with interpreting word-level context and meaning but which have yet to gain traction in mainstream AF research.

The four areas are: named entity recognition, summarization, semantics, and corpus linguistics methods. While space constraints limit discussion to these four areas, the list should not be viewed as exhaustive and should not prevent AF researchers exploring other techniques in the CL space, of which there are many.[20]

7.1 Named entity recognition

Named entity recognition (NER) is an information extraction task that isolates and then classifies named entities into predefined categories such as person names, locations and organizations. The simplest NER method for extracting named entities relies on handcrafted lists such as a gazetteer of all geographical locations, personal names or organization names. Rau (1991) applied a bag-of-words NER method using a set of handcrafted rules to detect whether a certain word refers to a company name and found that approximately four percent of tokens in a one million-word corpus of financial news are constituents of company names. The main problem with this approach is that it is hard to provide comprehensive coverage of all possible names.

A more appealing approach is a predictive system that uses statistical inference to detect patterns associated with a particular entity (Stevenson and Gaizauskas 2000). Accordingly, more recent NER work relies on statistical NLP methods developed for information extraction including Hidden Markov Models (HMMs) (Leek 1997), Conditional Markov Models (CMMs) (Borthwick 1999) and Conditional Random Fields (CRFs) (Lafferty et al. 2001). Well-known machine learning NER systems include the Stanford NER (Finkel et al. 2005) and the English NER (Sang and De Meulder Fien 2003). These methods rely on large volumes of manually annotated training data to achieve high accuracy. For example, the English NER dataset comprises manually annotated news stories between August 1996 and August 1997 from the Reuters Corpus. Retraining or reannotation is not usually required as these methods learn the information involved in extracting a certain named entity and are, therefore, able to extract names that do not appear in the original training corpus. Ratinov and Roth (2009) show that the NER systems trained on CONLL 2003 are domain independent.

---

[20] For example, Song (2018) applies a deep learning algorithm incorporating semantic-level features to business descriptions provided in Item 1 of firms' 10-K filings to construct a firm-level measure of industry disaggregation. Grüning (2011) uses unsupervised AI methods to construct a measure of annual report disclosure intensity. Chen et al. (2013) extract opinioned statements from 10-K filings using conditional random field techniques that reflect combinations of linguistic factors including morphology, orthography, predicate-argument structure, syntax and semantics.

Chen et al. (2013) and Hope et al. (2016) use the Stanford NER tool (http://nlp.stanford.edu:8080/ner/) in a financial context. Chen et al. (2013) extract person names from 10-K filings to support opinion mining of annual reports, while Hope et al. (2016) measure the specificity of firms' qualitative risk-factor disclosures (10-K, Item 1A) by extracting specific entity names belonging to the following entity categories: (1) names of persons, (2) names of locations, (3) names of organizations, (4) quantitative values in percentages, (5) money values in dollars, (6) times, and (7) dates. Other examples where the approach could be used productively include: detecting firms and individuals cited in SEC enforcement actions; extracting names of senior managers and directors to assist with a network analysis; identifying peer firms used in CEO relative performance evaluation arrangements; and retrieving the identities of corporate advisors (e.g., underwriters, compensation consultants, legal counsel, etc.).

7.2 Summarization

The volume of available information is increasing sharply and therefore the study of NLP methods that automatically summarize content has grown rapidly into a major research area. At the conceptual level, text summarization is the process of distilling content a single document or a set of related documents down to the most important events presented in the correct sequence. Automatic text summarization is therefore the process of producing a condensed version of a text using computerized methods. The aim is for the summary to convey the key contributions of the original text. Automated text summarization therefore involves identifying key sentences. The process of defining key sentences is highly dependent on the summarization method used.

Two generic approaches to automatic summarization are extractive (Funk et al. 2007, Bawakid and Oussalah 2008) and abstractive (Ganesan et al. 2010, Genest and Lapalme 2012). Extractive summarization methods are the dominant text summarization approach in NLP. The approach extracts key sentences or paragraphs (up to a specified limit) from the original text and then orders those sentences in a way that yields a coherent summary. Most summarizer systems operate at the sentence level, although extracted units can differ significantly depending on the specific summarization algorithm used. In contrast, abstractive summarization applies language dependent tools and natural language generation technology with the aim of mimicking human summarization methods. Accordingly, this summarization approach can yield words that are not present in the original document (i.e., paraphrasing rather than simply removing non-key content) and, therefore, tends to be more challenging to implement. Both extractive and abstractive summarization approaches have been used in single- and multi-document text summarization settings. The length of the generated summary can vary and ultimately depends on the purpose of the summarization process. The compression ratio (i.e., how much shorter the summary is than the original) varies from short summaries (e.g., 100 words) (Angheluta et al. 2004) to very short summaries (e.g., ten words) (Douzidia and Lapalme 2004).[21]

---

[21] Task descriptions of the summarization tracks organized by the Document Understanding Conference (DUC, http://duc.nist.gov/) and the Text Analysis Conference (TAC, https://tac.nist.gov/) specified the summary length to be between 240 and 250 words inclusive, white-space-delimited tokens. Summaries over the size limit were truncated and summaries below the size limit were penalized.

Extraction tends to play an important role in single- and multi-document summarization since the process works on extracting sections of a document or a collection of documents that convey the key contributions of the text.[22] Statistical metrics are normally used to assess the importance of extracted units. Early approaches to text summarization worked on allocating each unit a score based on features such as word frequencies (Luhn 1958), position in the text (Baxendale 1958), and the presence of key phrases (Edmundson 1969). The main drawback of these approaches is the difficulty of extracting semantically-related sentences because the methods do not capture linguistic features such as differentiating between nouns and verbs. These limitations motivated the adoption of more sophisticated approaches that are able to consider and extract such features. Machine learning is one approach that is used to identify important features, supported by NLP techniques for identifying key passages and relations between words and sentences (i.e., semantic similarity) (Kupiec et al. 1995, Jaqua et al. 2004, Leite and Rino 2008). Other methods rely on statistical approaches. For example, Fung and Ngai (2006) use Hidden Markov Models to reflect that the probability of including a given sentence in the summarization extract likely depends on whether the previous sentence has also been included. Clustering algorithms that assign observations into subsets (clusters) have also been used to group relevant sentences together to form a summary (Kruengkrai and Jaruskulchai 2003, Liu and Lindroos 2006). Statistical methods are useful when no training data are available but their main drawback is high sensitivity to outliers and noise (Herranz and Martinez 2009). The advantage of using machine learning approaches lies in the fact they can better mimic human summarization techniques. Consistent with other supervised machine learning applications, the main drawback of the approach is the cost of creating training data, which usually requires human participants to generate a large number of summaries (McCargar 2004).

---

[22] A key section may take the form of a sentence, paragraph or a number of n-grams.

Cardinaels et al. (2017) is the only AF study of which we are aware that uses statistical and heuristic summarizers to generate summaries of financial disclosures. Results reveal that automatic, algorithm-based summaries of earnings releases are generally less positively biased than management summaries, and that investors who receive an earnings release accompanied by an automatic summary arrive at more conservative (i.e., lower) valuation judgments. Other promising applications of the technology could include: detecting bias in IPO prospectuses, shareholder letters and other important communications with market participants; isolating statistically important sentences within a disclosure and then testing how management order and present that information as a means of either obfuscating or enhancing informativeness; and detecting bias in earnings announcements relative to annual report commentary in the spirit of Davies and Tama-Sweet (2012).

7.3 Semantics

Semantics is concerned with analysing the meaning of text and therefore underpins many aspects of NLP. For example, NER (section 7.1) focusses on the task of extracting meaning or classifying personal names, locations, organisations, etc., while summarisation (section 7.2) can be viewed as a form of semantic analysis that preserves the meaning of a document while reducing its length. Semantic analysis can occur at various levels including the entire document or corpus, a paragraph, a single sentence, or an individual word.

One of the primary sub-fields of semantics in the NLP domain is word sense disambiguation (WSD). The WSD task sounds deceptively simple: to correctly resolve the meaning of a word in a given context, such as distinguishing whether the word "above" refers to

Electronic copy available at: https://ssrn.com/abstract=3330757

performance against target or the physical location of content in a document. Complexity arises in WSD since we need an appropriate sense inventory against which to resolve the correct sense. Comparing the list of definitions of a word in two or more established dictionaries typically reveals differences even though the dictionaries are produced by lexicographers with expertise in resolving word meanings. WSD methods can be a mix of rule-based, probabilistic and more recently neural network-based approaches. Approaches tend to exploit the context of the word in question and the features that are important for disambiguation. WSD can be viewed as a sophisticated annotation task and often builds on POS tagging (section 3.2) since POS helps determine the correct meaning of a word (e.g., distinguishing "bank" as a verb from the noun sense). Meanwhile, using coarse-grained semantic fields that group senses together into broader categories (e.g. medicine, education, money and commerce) provides a way to tackle the sparse data problem and produce more accurate results (Rayson et al. 2004). (See Table 1 for details of semantic taggers.)

      Potential applications of semantics to AF research are endless. Arguably the simplest but potentially most profound opportunity is in the area of automated content analysis (i.e., unigram word lists) that characterizes the majority of work in the field (Figure 1, Panel B). Distinguishing the meaning of words to be counted from a simple list will reduce measurement error by isolating more precisely the specific concepts that the researcher is seeking to capture. Application of semantics therefore offers a natural extension to Loughran and McDonald's (2011) work on domain-specific lexicons by enabling identification of word senses from other contexts or domains outside AF that should not be counted when measuring content in financial narratives. Similarly, grouping words and phrases into related semantic fields provides a means of identifying topic areas in a corpus and then testing how these topic profiles vary across time,

firms, sectors, and countries. Accordingly, semantics provides a means of refining topic modelling approaches such as LDA where themes are identified purely on the basis of statistical words patterns that do not reflect meaning.

7.4 Corpus linguistics

Corpus linguistics is a methodological approach to discourse analysis that encompasses quantitative and qualitative techniques. It investigates discourse constructions by extracting important information about the nature and features of language based on empirical evidence. Corpus methods have been widely applied in the social sciences to discourses of gender (Baker 2014), refugees (Baker et al. 2008), grammar education (Conrad 2000), academic texts (Conrad 1996), and healthcare (Adolphs et al. 2004). Corpus linguistic analyses can be used to examine a wide variety of issues including: highlighting the evolution of a discourse type through time; discovering the linguistic differences and similarities among different parts of a document; comparing the structure and content of discourses provided in different languages; identifying the genre characteristics of a text type; and detecting specific characteristics associated with a text type including sentiment, thematic patterns, and important (key) words and themes. The methods are often used to triangulate extant studies that rely on researcher intuition (Baker 2006; Crawford and Csomay 2016). They may be used as an initial step in an NLP investigation to help understand the corpus and develop resources such as wordlists and topic categories for subsequent use; or the techniques can form the basis of a discourse analysis in their own right. The benefits of a corpus linguistic approach reflect the advantages of the CL method more generally insofar as it can help mitigate procedural researcher bias and uncover patterns in linguistic constructs that are hard to detect manually (Baker 2006: 13).

The typical corpus linguistics approach involves the following steps: selection and collection of data (section 3.1); data annotation and processing (section 3.2); and data analysis involving extraction of quantitative or qualitative information. Two important aspects of the corpus method are context analysis and keyness. Context analysis uses clustering and collocation tools to understand the meaning of words (or lexical items) by the company they keep (Firth 1957).[23] Specifically, clusters and collocates help to reveal how discourse is constructed around a particular lexical item. Clusters are common phrases containing a particular lexical item. These short phrases can be listed in order of frequency and used to pinpoint a lexical item's immediate and most frequent context. For example, identifying words that cluster frequently with "cost" could help distinguish cost reductions (good news) from cost increases (bad news). Collocation extends cluster analysis by identifying words that co-occur with a selected lexical item more often than is explained by chance (Baker 2006: 95). Collocation requires the researcher to preselect a word span (i.e., $\pm n$ words relative to the key lexical item) and, unlike clusters, collocates and key lexical items are not constrained to occur consecutively. Collocation has endless potential applications in AF, examples of which include developing a list of line items excluded from proforma earnings metrics or identifying points of view associated with a predefined topic such as Brexit.

Keyness provides a formal means of comparing a focal corpus against a reference corpus to distinguish lexical items that are either over-used or under-used in the focal corpus. Lexical items may be individual words (word keyness) or groups of words with a shared meaning (semantic keyness). Keyness provides insight into a discourse's defining constructs (Baker 2006:

---

[23] Although corpus linguists are primarily interested in lexical items (nouns, verbs, adverbs and adjectives), grammatical words have been shown to contribute to textual accessibility (see Epstein 2002). In this section, the term lexical item could mean lexical or grammatical words (including multi-word units) or groups of words with a shared meaning.

121, 125). The process involves compiling a list of lexical items that occur in the two corpora, documenting the frequency of each item in each corpus, and determining whether there is a statistically significant difference in the frequency of each item between the two corpora (normally assessed using a log likelihood test). Critically, keyness differs from simple frequency of occurrence: a word appearing very frequently in the focal corpus may not be classified as key if its expected incidence is also high based on the reference corpus, and vice versa.

Corpus linguistic methods have been used outside mainstream AF to study financial discourses including annual reports (Charteris-Black and Ennis 2001, Rutherford 2005, Lischinsky 2011a, Wang et al. 2012), letters from the board chair (Thomas 1997, de Groot et al. 2006), CEO letters (Hyland 1998, Conaway and Wardrope 2010, Dragsted 2014), CEO presentations (Rogers 2000), audit reports (Tian and Liang 2011), corporate press releases (Maat 2007), investment fund managers' reports (Bruce 2014) and corporate social responsibility reports (Lischinsky 2011b). A feature of this body of work is that it is often conducted by corpus linguists who may lack the level domain expertise required to address issues involving advanced economic theories and technical financial issues.

## 8. Conclusions

The justification for *not* using computational linguistics approaches to study unstructured financial market data centres on concerns over how the methods are applied and the relevance of the research questions examined as the most credible arguments against automated content analysis approaches. We review extant research in light of this concern using the following three lenses: core principles that underpin the CL endeavour; the advantages of CL methods (over manual coding) frequently cited by proponents and summarized by Li (2010a); and the relevance of CL methods and research agendas to policy and practice. Our review highlights concerns in all three areas and suggests that future work needs to proceed with caution to ensure proposed contributions are robust and meaningful.

We conclude that the main limitation of most prior AF work using automated content analysis methods is its heavy reliance on simple bag-of-words content analysis methods that fail to reflect context and disambiguate meaning. Extant research also clusters around a small number of text sources (primarily 10-K filings) and a limited set of discourse features such as tone and readability, raising concern that agendas are being shaped as much by ease of data access and implementation as they are by research priorities in financial discourse. A key theme emerging from our analysis is that CL methods and high-quality manual analysis represent complementary approaches to analyzing financial discourse (e.g., Black et al. 2018, Bentley et al 2018). Indeed, many CL applications require extensive manual input throughout the research pipeline including training, interpretation and evaluation. While there exists little doubt that adoption of CL methods represents a game-changing development for mainstream accounting and finance research, care is nevertheless required to avoid overstating its contribution while understating the value of alternative methodologies. The challenges for researchers choosing the CL option are first to ensure the research question aligns closely with the fundamental comparative advantages of scalability and latent feature detection, and then to ensure implementation follows best practice in CL rather than established practice in mainstream accounting and finance.

**Appendix A: Framework for computational linguistics analysis**

This appendix presents a simple framework and motivating examples designed to guide AF researchers in the basics of analysing financial discourse using CL methods. We disaggregate the CL pipeline into two core elements: collection and preparation of text for subsequent analysis (Figure A1), and methods and strategies for analysing a large dataset of financial text (Figure A2). Methods referenced in the following discussion are explained in more detail in the main body of the paper.

Figure A1 illustrates the three steps required to prepare a financial narratives dataset for further CL analysis. The c*orpus creation* step involves collection (either manually or automatically) of financial documents comprising either a single genre such as 10-K filings or multiple genres such annual reports, earning announcements and conference calls. Document genre reflects the style of language used in the discourse. Using a corpus of multiple genres creates the additional challenge of choosing the particular document types to include in the final dataset (Finn and Kushmerick 2003, Kothari et al. 2009).

The second step involves document *cleaning and pre-processing*. This typically involves choice of whether the entire document or specific parts of the document will be analysed. For example, analysis of the complete annual report narrative may be most appropriate for some applications whereas other research questions may be better suited to analysing specific sections such as the MD&A or the chair's letter. Where specific subsets of text are to be analysed, researchers need to specify how to identify and extract this content, and ideally provide evidence (i.e., evaluation) that the extraction process has retrieved the target content accurately. Accurate retrieval requires inclusion of all relevant content and exclusion of all irrelevant content.

The final step in the process of preparing a corpus for analysis involves deciding what (if any) *annotation* at apply. Annotations are used to identify the nature of words or other linguistic features such as sentences or phrases. This may involve manual, automated or a combination of manual plus automated analysis. Examples of manual annotation might include one or more coders defining the start and end of specific sections (El-Haj et al. 2019) or determining the tone of sentences (e.g., positive, negative or neutral) (Li 2010b). Automated annotation involves adding features to the text mechanically using an algorithm. Examples include identifying parts of speech (e.g., nouns, verbs and adjectives) and broader semantic categories (e.g., strategy, financial performance, and governance).

After applying steps 1-3 in Figure A1, the resulting corpus is ready for further analysis. This can involve using a simple approach such counting the frequency of words from a predefined list (dictionary) or applying more sophisticated CL methods such as pre-trained machine learning classifiers. In addition, sub-corpora may be created based on metadata such as time, sector and firm size for comparison using corpus linguistics techniques such as keyness. Regardless of the specific approach(es) adopted, a fundamental goal is to ensure the analysis reflects accurately the meaning of discourse(s) in the corpus. Figure A2 summarizes the processes involved in performing a range of CL analysis tasks on a corpus of financial documents. Figure A2 also highlights structural differences and drawbacks associated with simple content analysis approaches typically adopted in AF research compared with the more sophisticated techniques often applied in CL. The primary concern with approaches to textual analysis currently favoured by AF researchers is whether they are sufficiently powerful to capture correct meaning in the underlying text. This concern is particularly germane to AF given the high degree of specialist (domain-specific) language involved, combined with the subtleties that characterize many aspects of the financial communication process. The following two example tasks illustrate the type of choices that researchers face.

**Example 1:** The task of detecting human personal names (e.g., members of the board of directors) can be approached using either a dictionary of names or a pre-trained Named Entity Recognition (NER) tool to automatically perform the task. Major drawbacks associated with the dictionary approach include its static nature (there is no scope for "learning" to reflect structural changes in, say, gender balance), incompleteness (it is very difficult to cover all feasible names), and the failure to differentiate between locations and personal names (e.g. Stuart Lancaster or Johnny English). On the other hand, care is required when using pre-trained NER approach to ensure the domain used to train NER system accurately reflects language norms in the corpus where it is being applied. For example, an NER system trained on detecting English names found in British newspapers may perform poorly if used to detect board member names in German annual reports.

**Example 2:** Consider an AF researcher wanting to determine sentence-level tone in the management discussion and analysis (MD&A) section of U.S. registrants' 10-K filings. The first task in the CL pipeline involves creating a corpus following the steps summarized in Figure A1. This can be achieved by downloading a sample of 10-Ks manually from EDGAR or alternatively using EDGAR's API to automatically harvest filings. The latter approach is faster and provides the opportunity to collect more reports, thereby increasing sample size. However, the manual approach may be more precise, particularly where the retrieval task involves collecting filings for a specific set of firms such as those subject to Securities and Exchange Commission Enforcement Actions.

The cleaning and pre-processing step illustrated in Figure A1 involves retaining only the content relating to Item 7 (MD&A) for further analysis. This step can be implemented manually or automatically using a script. Note, however, that an automated approach may still involve aspects of manual annotation. For example, manual annotation may be used to delineate more precisely the beginning and end of the MD&A section to support retrieval where the sample of 10-Ks is small, or to generate a gold standard dataset for validation purposes.[24] Whether or not the researcher elects to apply further manual or automated annotation to the cleaned corpus of MD&A commentary will depend to a large degree on the specific research question examined and the particular CL method used to test the question. The final stage in Figure A1 is therefore heavily dependent on the particular CL strategy adopted from Figure A2.

Developing an automated process for measuring MD&A sentence-level tone requires the researcher to select various processes from Figure A2. The first step involves splitting MD&As into sentences. This may be achieved using certain delimiters (e.g., a full stop) to determine the end of each sentence. Alternatively, a pre-trained sentence splitter may be applied. Using delimiters such as full stops may be error prone because the delimiter may also appear in titles (e.g. Dr. Johns) or numbers (e.g. 1.5 million), resulting in incomplete sentences and increasing measurement error. Pre-trained tools such as Stanford Sentence Splitter are able to automatically detect cases such as "Dr. Johns" and "1.5 million", thereby minimizing incorrect text splits.[25]

With the MD&A text split into sentences, the researcher must choose the approach to measuring tone. One approach is to use a pre-trained sentiment analyser such as the Stanford sentiment tool.[26] Using pre-trained (off-the-shelf) tools helps reduce time and effort. However, caution is required if the tool has been trained in a domain where the language and writing style differ from the 10-K. Stanford's sentiment analyser was trained on U.S. movie reviews. Since the language of movie reviews is likely to differ considerably from the typical 10-K discourse, sentiment results may be unreliable despite the cutting-edge nature of Stanford method. Instead, the domain-specific nature of 10-K language may require development of a new classifier trained on a relevant financial narratives dataset with sentences manually tagged as positive, negative or neutral (e.g., Li 2010, El-Haj et al. 2016).

---

[24] In the case of 10-Ks, the annotation process is relatively straightforward due to the structured nature of the 10-K filing template and the availability of filing data in HTML file format.
[25] https://nlp.stanford.edu/software/tokenizer.shtml
[26] https://nlp.stanford.edu/sentiment/code.html

Alternatively, the researcher may prefer to measure tone using word-frequency counts based on dictionaries of positive and negative words. Dictionaries may be derived from general language (e.g., Harvard-IV-4) or domain-specific discourse (e.g. Loughran and McDonald 2011). While word-frequency counts are the dominant approach used in AF, the risk of false positive and false negative results is high due, for example, to the common occurrence of negation in English language. For instance, a sentence such as "The Group's net loss decreased by $13.6 million from $17.2 million in 2009" is likely to be classified as negative due to the presence of words such as loss and decreased, even though the sentence is actually positive.

Adding further annotations to the MD&A corpus could help reduce measurement error associated with both the machine learning classifier and dictionary approaches. For example, both approaches may benefit from information on additional text features such as knowing the part of speech (POS) of each word in a sentence. Automated POS annotation is possible using a pre-trained machine learning algorithm such as Stanford's Part of Speech (POS) tagger.[27] Other levels of language annotation that can help reflect meaning in a sentence and therefore enhance tone classification performance include grammatical tagging and semantic tagging (see Table 1).

---

[27] https://nlp.stanford.edu/software/tagger.shtml

62Electronic copy available at: https://ssrn.com/abstract=3330757

While such approaches are commonplace in CL, few studies in AF currently adopt these methods.

**Figure A1**: Steps involved in preparing a financial narratives dataset for further CL analysis

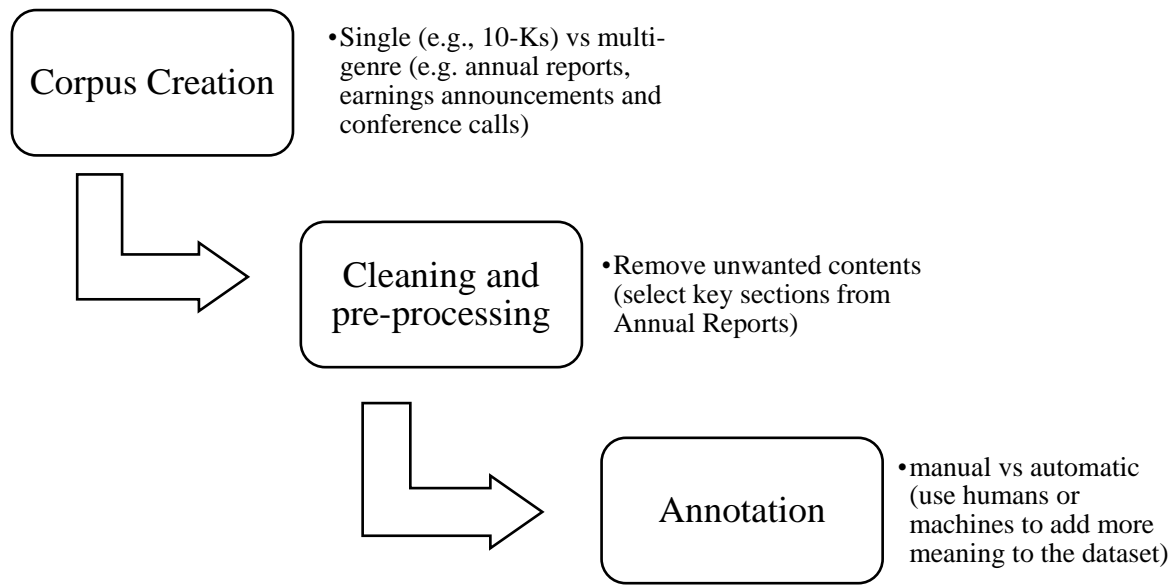

**Figure A2:** Summary of corpus analysis methods and strategies

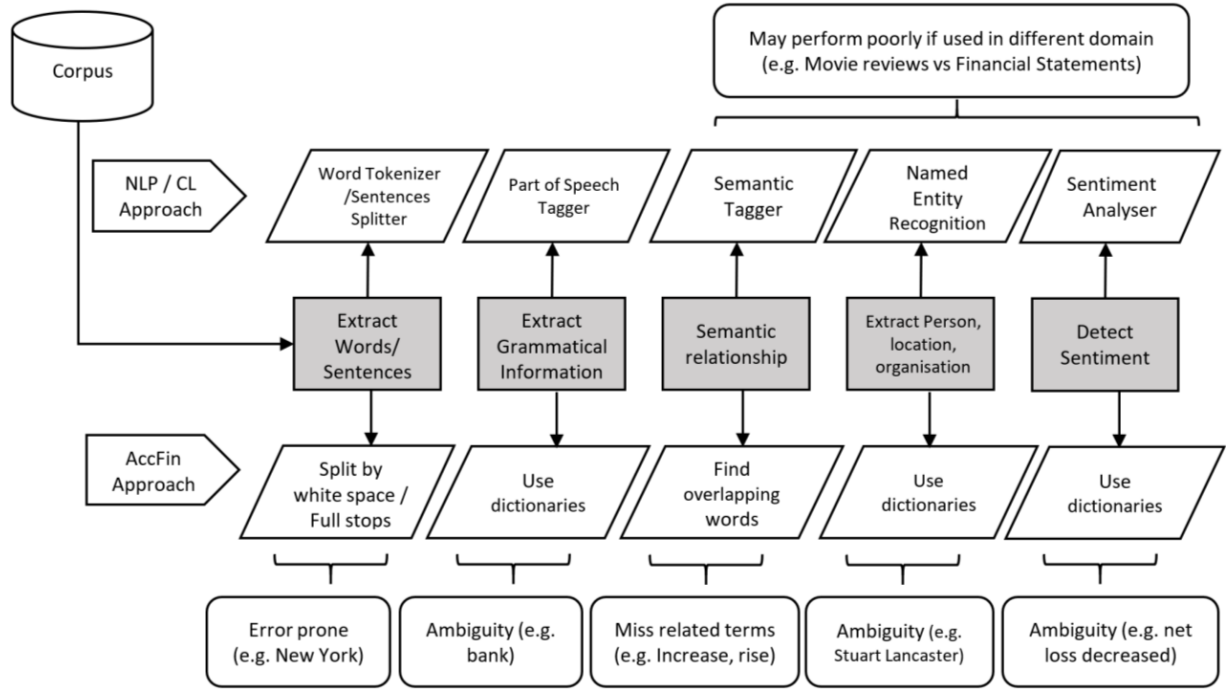

**Appendix B: Does the Fog index reflect financial reporting quality?**

The Fog index is frequently used in accounting and finance (AF) research to proxy for quality attributes of financial discourse including clarity of presentation (i.e., understandability) and degree of managerial obfuscation (i.e., understandability). The measure captures two primitive surface-level discourse features: sentence length and word complexity. Numerous published papers report evidence consistent with theoretical predictions concerning the determinants and consequences of financial discourse quality using Fog. However, Loughran and McDonald (2014b, 2016) and Bonsall et al. (2017) highlight structural problems with the Fog measure that also apply to its close relations such as Flesch Reading Ease, Flesch-Kincaid Grade Level, SMOG Index, Automated Readability Index, and Linsear Write Formula. In the absence of manual evaluation of the validity of Fog as a measure financial discourse quality, we view its reliability as unresolved question that warrants further investigation.

To highlight reliability concerns we report results of a simple manual evaluation exercise designed to provide illustrative (rather than definitive) evidence on the degree to which Fog correlates with the quality of annual report commentary as evaluated by domain experts. We focus on annual reports due to their popularity as a discourse source in prior research (Figure 1, Panel B). The starting point for our analysis is the set of U.K. annual reports published between 2007 and 2014 by FTSE-350 firms that are identified by domain experts as exemplars of best reporting practice in a given fiscal year. A range of organizations rate annual reports published by large U.K.-listed firms including PwC, Investor Relations (IR) Society, ICSA Hermes, Communicate Magazine, Accountancy Age, and Report Watch. We pool annual report winners identified by these organizations to form an initial sample of exogenously-rated reports considered by accounting experts to represent the best examples of high quality reporting practice. Each award winning report is then matched with a non-winning report published in the same fiscal year by a FTSE-350 firm operating in the same sector and with the closest propensity score (1:1 nearest neighbour matching with replacement and a 0.068 calliper) derived from a model that includes the following covariates: firm size, analyst following, earnings variability, return on equity, leverage, auditor, and a suite of governance variables.

Thirty matched pairs are selected randomly for use in our subsequent analysis. Fog scores are computed for selected winner and non-winner pairs using the Fathom algorithm (Svoboda 2013; https://github.com/ogrodnek/java_fathom). Scores are derived using all content for the narrative component of the annual report, which in the U.K. typically includes the chair's letter to shareholders, CEO review, CFO review, strategic report, directors' report, corporate governance statement, remuneration report, report on principal risks and uncertainties, and sundry sections such as corporate social responsibility disclosures.

Report pairs are also rated independently for quality by 37 participants attending a workshop on textual analysis methods held at Lancaster University in September 2018. Participants comprised a mixture of faculty and PhD students who were either researching financial discourses or who had general experience of teaching and researching financial reporting. Each participant received reports for two winner-non-winner pairs (four reports in total per participant), resulting in 74 paired scores (148 individual report scores) in total. Report pairs were allocated to a least two participants to facilitate calculation of an average rating for each report to reduce the sensitivity of results to personal perceptions of quality. Reports were provided to participants prior to the workshop without any contextual information and no indication of pairwise matching or report quality. Participants were requested to rate the narrative component of each report on the basis of overall clarity and understandability using a five-point scale, with five indicating clear presentation that is straightforward to understand and one indicating low clarity and poor understandability. Participants were directed to provide absolute ratings rather than relative rankings. Scores were submitted prior to attending the workshop.

Table B1 provides Fog and participant quality ratings. Panel A presents report-level averages for winner-non-winner matched pairs. Columns 2-4 report Fog scores for the winner and non-winner samples. The mean (median) for both samples is around 20.7, consistent with prior evidence suggesting that 10-K annual reports are challenging documents to read. Paired parametric and nonparametric tests reveal no statistical difference in Fog scores across the two groups. The results do not support the view that Fog captures reporting quality as perceived by organizations that specialize in financial reporting and investor communication.

Columns 5-7 in Panel A present workshop participants' average blind ratings of report quality. Unlike results for Fog, mean and median participant ratings are statistically higher for the winner sample at the 0.06 level or better, suggesting that annual report quality is an observable phenomenon with common components evident to different parties. Columns 8-10 report results for a binary transformation of quality ratings where reports with scores above three are treated as high quality and allocated a value of one, while reports scoring three and below are treated as non-high quality and allocated a value of zero. Results confirm that winner reports are judged by participants to be of higher quality. Collectively, evidence in Panel A suggests that domain experts are able to distinguish report quality whereas Fog displays no meaningful discriminatory power. Results also highlight considerable variation in participants' individual quality ratings, consistent with the view that annual report quality is a challenging construct for human coders to define and measure.

Panel B of Table B1 reports Fog scores by participants' individual report ratings, where high quality is defined as a rating above three. Only 64 of the 148 ratings provided ranked a report as high quality. Evidence that individuals' personal quality ratings do not map directly on to professional organizations' quality rankings is provides further support for the view that reporting quality is a slippery construct to measure. Similar to findings reported in Panel A, Fog continues to lack discriminatory power for this alternative quality ranking.

Results from this illustrative exercise cannot be used to draw definitive conclusions about the reliability or otherwise of using readability scores generally, and Fog scores in particular, as a proxy for the quality of annual report discourse. At a minimum, however, the findings serve to illustrate the challenges associated with using a naïve bag-of-words approach to capture such a poorly-specified construct as the clarity, understandability and overall quality of a complex document such as an annual report.

**Table B1**: Summary statistics for annual report quality

*Panel A*: Scores by award winner and non-winner categories

|  | Fog | | | Participant quality rating (raw) | | | Participant quality rating (high) | | |
| --- | --- | --- | --- | --- | --- | --- | --- | --- | --- |
|  | Winner | Non-winner | p-value for paired difference | Winner | Non-winner | p-value for paired difference | Winner | Non-winner | p-value for paired difference |
| Mean | 20.620 | 20.914 | 0.308 | 3.506 | 3.111 | 0.040 | 0.544 | 0.339 | 0.017 |
| St. dev | 1.330 | 0.957 |  | 0.560 | 0.735 |  | 0.300 | 0.298 |  |
| Median | 20.552 | 20.859 | 0.580 | 3.500 | 3.333 | 0.054 | 0.500 | 0.333 | 0.021 |
| N | 30 | 30 |  | 30 | 30 |  | 30 | 30 |  |

*Panel B*: Fog by participant rating

|  | Participant quality rating: | | p-value for difference |
| --- | --- | --- | --- |
|  | High | Non-high |  |
| Mean | 20.601 | 20.813 | 0.284 |
| St. dev | 1.254 | 1.137 |  |
| Median | 20.471 | 20.929 | 0.299 |
| N | 64 | 84 |  |

This table reports findings for alternative measures of annual report quality. The sample comprises 30 award winning annual reports published by FTSE-350 firms between 2007 and 2014. Awarding bodies are PwC, Investor Relations (IR) Society, ICSA Hermes, Communicate Magazine, Accountancy Age, and Report Watch. Each award winning report is matched with a non-winning report published in the same fiscal year by a FTSE-350 firm operating in the same sector and with the closest propensity score derived from a model that includes the following covariates: firm size, analyst following, earnings variability, return on equity, leverage, auditor, and a suite of governance variables. The resulting 30 matched pairs are independently blind-rated by 37 participants at an accounting research workshop. Participants rate the narrative component of each report on the basis of overall clarity and understandability using a five-point scale, with five indicating clear presentation that is straightforward to understand and one indicating low clarity and poor understandability. Each participant received reports for two winner-non-winner pairs (four reports in total per participant), resulting in 74 paired scores (148 individual report scores) in total. Report pairs were allocated to a least two participants to facilitate calculation of an average rating for each report to reduce the sensitivity of results to personal perceptions of quality. Panel A presents report-level average ratings together with the Fog index computed for winner and non-winner pairs using the Fathom algorithm (Svoboda 2013). Fog ccores are derived using all content for the narrative component of the annual report, which in the U.K. typically includes the chair's letter to shareholders, CEO review, CFO review, strategic report, directors' report, corporate governance statement, remuneration report, report on principal risks and uncertainties, and sundry sections such as corporate social responsibility disclosures. High quality reports in columns 8-10 are those with a raw score greater than three. Probability values relate to two-tailed paired t-tests (means) and paired Wilcoxon Signed Rank tests (medians). Panel B reports Fog scores by workshop participants' report level quality rating. Probability values relate to two-tailed two-sample t-tests (means) and Wilcoxon Signed Rank tests (medians).

Fisher, I., Garnsey, M., Hughes, M. (2016). Natural language processing in accounting, auditing and finance: A synthesis of the literature with a roadmap for future research. *Intelligent Systems in Accounting, Finance and Management* 23(3): 157-214

Frankel, R., Jennings, J., Lee, J. (2016). Using unstructured and qualitative disclosures to explain accruals. *Journal of Accounting and Economics* 62(2-3): 209-227

Francis W. N., Kucera H. (1979). Brown Corpus manual: Manual of information to accompany a standard corpus of present-day edited American English, for use with Digital Computers. http://clu.uni.no/icame/manuals/BROWN/INDEX.HTM (accessed 14 March 2019)

Francis, J., Schipper, K., Vincent, L. (2002). Expanded accounting disclosures and the increased usefulness and earnings announcements. *The Accounting Review* 77(3): 515-546

Fung P., Ngai G. (2006). One story, one flow: Hidden Markov story models for multilingual multidocument summarization. *ACM Transactions on Speech and Language Processing (TSLP)*. 3(2): 1–16

Funk A., Maynard D., Saggion H., Bontcheva K. (2007). Ontological integration of information extracted from multiple sources. The Multi-source Multi- lingual Information Extraction and Summarization (MMIES) workshop at Recent Advances in Natural Language Processing (RANLP07). Borovets, Bulgaria: RANLP07

Ganesan K., Zhai C. X., Han J. (2010). Opinosis: A graph-based approach to abstractive summarization of highly redundant opinions. In proceedings of the 23rd international conference on computational linguistics. 340–348

Garside R., Leech G., McEnery T. (1997). *Corpus Annotation: Linguistic Information from Computer Text Corpora*. Taylor and Francis

Garside, R., Smith, N. (1997). A hybrid grammatical tagger: CLAWS4, in Garside, R., Leech, G., McEnery, A. (eds.) *Corpus Annotation: Linguistic Information from Computer Text Corpora*. Longman, London: 102-121

Gao, Q., Lin, M. (2015). Lemon or cherry? The value of texts in debt crowdfunding. Unpublished working paper

Genest P.-E., Lapalme G. (2012). Fully abstractive approach to guided summarization. In proceedings of the 50th Annual Meeting of the Association for Computational Linguistics: Short Papers 2: 354–358
78Electronic copy available at: https://ssrn.com/abstract=3330757

International Integrated Reporting Council (2011). *Toward Integrated Reporting. Communicating Value in the 21st Century*. International Integrated Reporting Council. https://integratedreporting.org/wp-content/uploads/2012/06/Discussion-Paper-Summary1.pdf

Iyyer M., Manjunatha, V., Boyd-Graber, J., Daumé III, H. (2015). Deep unordered composition rivals syntactic methods for text classification. In proceedings of the 53rd Annual Meeting of the Association for Computational Linguistics and the 7th International Joint Conference on Natural Language Processing (Volume 1: Long Papers). 1. 1681–1691

Jagadeesh N., Wu, D. (2013). Word power: A new approach for content analysis. *Journal of Financial Economics* 110(3): 712-729

Jaqua L., Jaoua M., Belguith H., Ben H. (2004). Summarization at LARIS laboratory. In proceedings of the 4th Document Understanding Conferences

Jones, M. J., Shoemaker, P.A. (1994). Accounting narratives: A review of empirical studies of content and readability. *Journal of Accounting Literature* 13: 142–84

Junker, M., Hoch, R., Dengel, A. (1996). On the evaluation of document analysis components by recall, precision, and accuracy, Fifth International Conference on Document Analysis and Recognition (ICDAR)

Kaplan, A. (1964). *The Conduct of Inquiry: Methodology for Behavioral Science.* Transaction Publishers

Kearney, C., Liu, S. (2014). Textual sentiment in finance: A survey of methods and models. *International Review of Financial Analysis* 33: 171-185

Keusch, T., Bollen, L., Hassink, H. (2012). Self-serving bias in annual report narratives: An empirical analysis of the impact of economic crises. *European Accounting Review* 21(3): 623–648

Kim, J., Kim, Y., Zhou, J. (2017). Languages and earnings management. *Journal of Accounting and Economics* 63(2-3): 288-306

Kimbrough, M.D., Wang, I.Y. (2014). Are seemingly self-serving attributions in earnings press releases plausible? Empirical evidence. *The Accounting Review* 89(2): 635–667

Koller, V. (2011) *'Hard-working, team-oriented individuals': constructing professional identities in corporate mission statements.* In: Constructing Identities at Work. Palgrave Macmillan, Basingstoke81Electronic copy available at: https://ssrn.com/abstract=3330757

**Table 1**: Summary of automated annotation methods and resources

| Language feature | Description | Resource |
|---|---|---|
| Part of Speech (POS) | Classifies each word in a corpus according to its major word class (e.g. noun, verb, adjective, etc.), along with additional information (e.g. singular or plural, common or proper noun, etc.). Along with many other tasks, POS tags assist with word sense disambiguation and therefore help uncover meaning. See Garside and Smith (1997) for further details. | http://ucrel.lancs.ac.uk/claws/ <br> https://nlp.stanford.edu/software/tagger.shtml |
| Morphology | Examines the structure of words and parts of words (e.g., stems, root words, prefixes, and suffixes) in a corpus, and as well as POS, intonation and other features that provide insights into meaning. Studying word properties, how they are formed, and their relation to other words in the same language facilitates word and multiword sense disambiguation. See (Hajič 2000) for further details. | http://nlp.stanford.edu:8080/parser/ |
| Grammar | The set of rules that describe the structure of a language and control the way that sentences are formed to avoid producing unacceptable sentences (e.g., ambiguous, hard to understand, etc.). Grammatical analysis involves examining the form and arrangement of words, phrases, and sentences. Understanding word arrangement sheds light on the meaning of words and phrases. See Garside and Smith (1997) for further details. | http://nlp.stanford.edu:8080/parser/ |
| Syntax | One of the three levels of grammar, involving the grammatical arrangement of words and phrases (i.e., word order) to create well-formed sentences in a language. Syntax analysis, also known as parsing, is often conducted as a way to understand the exact meaning of a word or sentence (by helping to reduce ambiguity). The process decomposes a sentence or other string of words into its constituents and shows their syntactic relation to each other (e.g., subject, verb. object). For example, parsing the phrase "man eats burger" reveals that the singular noun "man" is the subject of the sentence, the verb 'eats' is the third person singular of the present tense of the verb 'to eat', and the singular noun 'burger' is the object of the sentence. See Danqi and Manning. (2014) for further details. | http://nlp.stanford.edu:8080/parser/ |
| Semantics | The process of extracting meaning from an individual word or a multi-word expression (MWE). For example, the phrase "I parked by the bank" has multiple meanings depending on the definition of the word 'bank'. A semantic tagger assigns meaning (a sematic category) to a word or MWE based in its most likely use in the discourse. The UCREL semantic analysis system uses a multi-tier structure with the tagset arranged in a hierarchy comprising 21 major discourse fields that expand to 232 finer-grained category labels. The UCREL tagger is available for multiple languages. See Rayson et al. (2004) for further details. Other forms of semantic analysis include Named Entity Recognition, summarisation, semantic role labelling and sentiment analysis. | http://ucrel.lancs.ac.uk/usas/ |

| Pragmatics | Concerns the way(s) in which context contributes to meaning. Whereas semantics examines "conventional" meaning as determined by language rules regarding grammar, lexicon, etc., pragmatics recognizes that transmission of meaning also depends on the context of the utterance including pre-existing knowledge about message sender and receiver, and the inferred intent of the message sender. Pragmatic analysis helps overcome ambiguity by relating meaning to the manner, place, time, etc. of an utterance. See Culpeper et al. (2008) for further details. | https://aclanthology.info/papers/L14-1287/l14-1287  (although manual annotation is more usual) |
|---|---|---|
| Discourse | Analysis of the patterns of language in texts that are longer than a single sentence. Discourse analysts are interested in the wider discourse context in order to understand how it affects the meaning of individual sentences and words. The approach aims at revealing socio-psychological characteristics of a person or persons. Framing forms an important part of discourse analysis. | Automated tagging for discourse levels are not well developed or widely used; manual annotation dominates in this area. See Prasad et al. (2007) for an example. |

**Figure 1**: Distribution of accounting and finance papers using computational linguistics methods.
*Panel A*: Distribution of papers by computational linguistics method

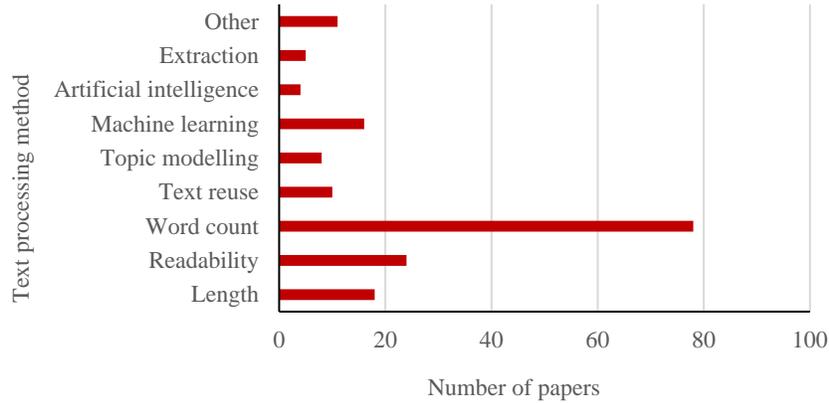

*Panel B*: Distribution of papers by text source

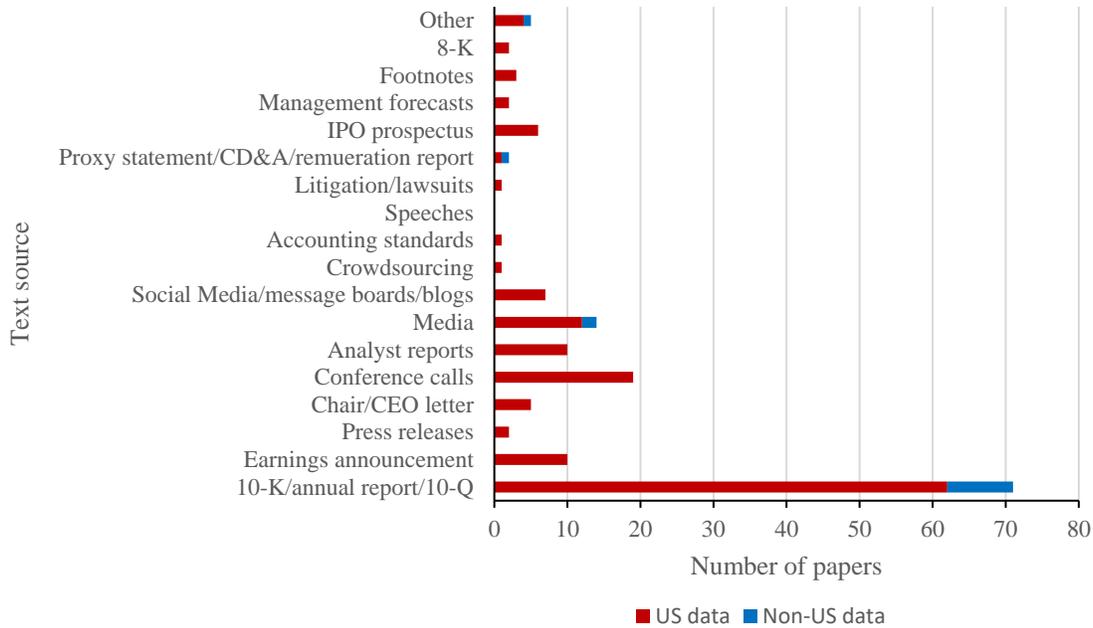

*Panel C:* Distribution of papers using keyword counts by text construct

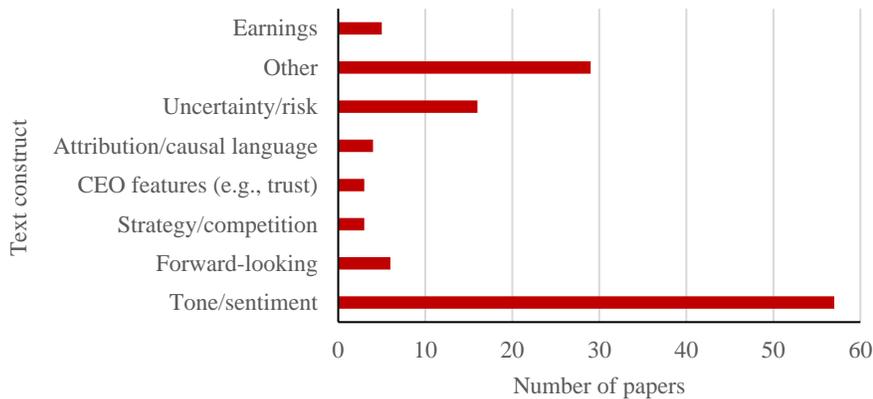

**Figure 1** *(continued)*

This figure summarizes the key features of 149 accounting and finance papers that apply a significant element of computational linguistic analysis, broadly defined. The sample includes published articles and working papers appearing since 2004. Criteria for inclusion in the sample are: (1) the study is published in a mainstream accounting and finance journal (or a journal with a distinct accounting and finance track) appearing on the tenure lists of international business schools; or (2) the study is a working paper available on the Social Science Research Network with a least 10 downloads and at least one co-author from an accounting and finance department in a research intensive business school. The complete list of 149 papers is available as an online appendix and at http://ucrel.lancs.ac.uk/cfie/. Panel A categorizes papers according to the computational linguistic method(s) applied, with multiple methods possible for an individual paper. Panel B classifies papers by text source(s), with multiple sources possible for a single paper. Panel C groups the subsample of studies that derive at least one linguistic construct using a wordlist by the particular constructs examined, with multiple constructs permitted per paper.